\documentclass[letterpaper]{article} % DO NOT CHANGE THIS
\usepackage{aaai23-r2hcai}  % DO NOT CHANGE THIS
\usepackage{times}  % DO NOT CHANGE THIS
\usepackage{helvet}  % DO NOT CHANGE THIS
\usepackage{courier}  % DO NOT CHANGE THIS
\usepackage[hyphens]{url}  % DO NOT CHANGE THIS
\usepackage{graphicx} % DO NOT CHANGE THIS
\usepackage{natbib}  % DO NOT CHANGE THIS AND DO NOT ADD ANY OPTIONS TO IT
\usepackage{caption} % DO NOT CHANGE THIS AND DO NOT ADD ANY OPTIONS TO IT
\usepackage{algorithm}
\usepackage{algorithmic}

\usepackage{newfloat}
\usepackage{listings}
\DeclareCaptionStyle{ruled}{labelfont=normalfont,labelsep=colon,strut=off} % DO NOT CHANGE THIS
\floatstyle{ruled}
\newfloat{listing}{tb}{lst}{}
\floatname{listing}{Listing}
\usepackage{array}
\usepackage{multirow}
\usepackage{pifont}
\usepackage{mathtools}
\usepackage{amsfonts}
\usepackage{textcomp}

\usepackage{lscape}
\usepackage{longtable}

\usepackage{fancyhdr}
\fancypagestyle{firstpagehf}
{
   \fancyhf{}
   \fancyhead[HC]{The AAAI 2023 Workshop on Representation Learning for Responsible Human-Centric AI (R$^2$HCAI)}
   \fancyfoot[C]{\thepage}
}

\title{Towards a Deeper Understanding of Concept Bottleneck Models Through End-to-End Explanation}

\author {
    Jack Furby\textsuperscript{\rm 1} \quad
    Daniel Cunnington\textsuperscript{\rm 2} \quad
    Dave Braines\textsuperscript{\rm 2} \quad
    Alun Preece\textsuperscript{\rm 1}
}
\affiliations {
    \textsuperscript{\rm 1} Cardiff University, UK \quad
    \textsuperscript{\rm 2} IBM Research Europe\\
    furbyjl@cardiff.ac.uk
}

\begin{document}
\thispagestyle{firstpagehf}
\maketitle

\begin{abstract}
   Concept Bottleneck Models (CBMs) first map raw input(s) to a vector of human-defined concepts, before using this vector to predict a final classification. We might therefore expect CBMs capable of predicting concepts based on distinct regions of an input. In doing so, this would support human interpretation when generating explanations of the model's outputs to visualise input features corresponding to concepts. The contribution of this paper is threefold: Firstly, we expand on existing literature by looking at relevance both from the input to the concept vector, confirming that relevance is distributed among the input features, and from the concept vector to the final classification where, for the most part, the final classification is made using concepts predicted as present. Secondly, we report a quantitative evaluation to measure the distance between the maximum input feature relevance and the ground truth location; we perform this with the techniques, Layer-wise Relevance Propagation (LRP), Integrated Gradients (IG) and a baseline gradient approach, finding LRP has a lower average distance than IG. Thirdly, we propose using the proportion of relevance as a measurement for explaining concept importance.
\end{abstract}

\section{Introduction} \label{Introduction_section}

Humans build mental models which are internal representations of an object's internal mechanics. These are used to predict an object's future states and aid in interactions \citep{10.5555/7909, The_nature_of_explanation_1943, 10.1145/800045.801613, Norman1983SomeOO}. If a human is unable to build an accurate representation of a Deep Neural Network (DNN) decision boundaries, the human could be misled to either accept misclassifications or disregard its output entirely. Explainable artificial intelligence (XAI) techniques aim to aid humans in building mental models of DNNs. One XAI technique is to use \textit{saliency maps}, a tool which highlights features of an input which were relevant to a prediction.

Concept Bottleneck Models (CBMs) \citep{pmlr-v119-koh20a} seek to enable richer human-machine interaction by training a DNN to predict a vector of human-defined concepts before using this vector to predict a final classification. CBMs have the potential of performing a task in a similar way to humans, whilst also enabling interpretability of their learned representations. For instance, to identify a bird, a human may first recognise parts of the bird such as the colour and size, before using this information for bird identification. As the final classification uses the predicted concept vector, a human user will be able to modify the concept vector, referred to as \textit{intervening}, and add or remove concepts to inspect changes to the final classification. %This enables the human to answer what-if questions, e.g., ``What if the model instead predicted these concepts?''.

Despite the concept vector output, CBMs are unable to explain what input features lead to concept predictions, or which concepts contribute to the final classification. An XAI study for CBMs \citep{margeloiu2021concept} used saliency maps and suggests that CBMs do not learn concepts as humans would expect, but instead attribute relevance to the entire input, and not to distinct regions. The authors however only looked at saliency maps for concepts and not the final classification and did not indicate what the models may be learning for concept predictions. In this paper we position XAI, and in particular saliency maps, as a technique to present the relevancy behind a CBM prediction, both concept and final classification and thus make the model reasoning accessible to a human.

Our research focuses on producing explanations for CBMs targeted for use by domain experts. For this reason, we use CBMs with Layer-wise Relevance Propagation (LRP) \citep{10.1371/journal.pone.0130140} to explore relevancy attributed, from the final classification to the concept vector and from the concept vector to the input as relevance can be attributed to groups of input features instead of on a pixel-by-pixel basis \citep{9369420}. Despite the primary use of LRP in this paper, the techniques described apply to other gradient-based attribution methods such as Integrated Gradients (IG) \citep{10.5555/3305890.3306024} which we also used to compare concept vector to input saliency maps. Our code is publicly available\footnote{Code: https://github.com/JackFurby/explainable-concept-bottleneck-models}.

\section{Background} \label{background_section}

\subsection{Concept Bottleneck Models}

CBMs can be given in the form $f(g(x))$ where the function $g$ refers to the prediction of concepts $\hat{c}$ using the input $x$ and the function $f$ is the prediction of the final classification $y$ with the input $\hat{c}$.

Given the training set $\{x^{(i)}, y^{(i)}, c^{(i)}\}_{i=1}^n$ where we are provided with a set of inputs $x \in \mathbb{R}^d$, corresponding targets $y \in \mathcal{Y}$ and vectors of $k$ concepts $c \in \mathbb{R}^k$. A CBM maps the input space to the concept space $g : \mathbb{R}^d \rightarrow \mathbb{R}^k$ and maps concepts to final targets $f : \mathbb{R}^k \rightarrow \mathcal{Y}$. This is such that the final classification is made using only concept predictions. 

CBMs can be trained in three ways: \textit{independent}, \textit{sequential} and \textit{joint}. With independent training, each model part is trained separately, whereas the sequential method trains the model parts one after another and joint trains them together in an end-to-end fashion.

In this paper, we refer to the model part predicting $c$ as $x \rightarrow c$ and the model part predicting $y$ as $c \rightarrow y$.

\subsection{Layer-wise Relevance Propagation}

LRP \citep{10.1371/journal.pone.0130140} is an explanation technique that propagates a prediction backwards through a network until a defined layer or input is reached, following a set of rules. Primarily, the network output is conserved and is only redistributed to the neurons of the previous layer. The total value propagated does not change. The use of alternative rules adds flexibility for LRP implementation \citep{Montavon2019}. These include the basic rule (LRP-0), Epsilon rule (LRP-$\epsilon$) and Alpha Beta Rule (LRP-$\alpha\beta$) \citep{10.1371/journal.pone.0130140}. Rules can be applied in a composite manner to overcome the shortcomings of any single rule.

LRP is considered to have an advantage over the related IG method \citep{10.5555/3305890.3306024} in that IG tends to produce very fine-grained pixel-wise mappings whereas LRP tends to group relevance to features from the input \citep{9369420}. As we are interested in mapping relevance attributed to concepts, and concepts occupy distinct regions in input images, this makes LRP an appealing choice. %LRP has also been shown to help human participants learn input features a model was sensitive to \citep{10.1145/3377325.3377519}.

\section{Setup} \label{setup_section}

\subsection{Models and Dataset}

\begin{figure*}[ht!]

\begin{center}
\begin{tabular}{m{0.13\textwidth} b{0.05\textwidth} m{0.13\textwidth} @{\hspace{0.5\tabcolsep}} m{0.13\textwidth} @{\hspace{0.5\tabcolsep}} m{0.13\textwidth} @{\hspace{0.5\tabcolsep}} m{0.13\textwidth} @{\hspace{0.5\tabcolsep}} m{0.13\textwidth} }
    \vspace{12.5mm}input &
    &
    has\_crown\newline\_color\newline::brown &  
    has\_wing\newline\_shape\newline::pointed-wings &  
    has\_back\newline\_color\newline::brown &  
    has\_bill\newline\_shape\newline::all-purpose &  
    has\_breast\newline\_pattern\newline::solid 
 \\
    \frame{\includegraphics[width=0.13\textwidth]{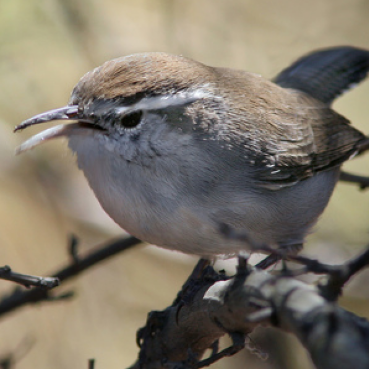}} &  
    \rotatebox[origin=b]{90}{\parbox{0.05\textwidth}{Independent and\\Sequential}} &
    \frame{\includegraphics[width=0.13\textwidth]{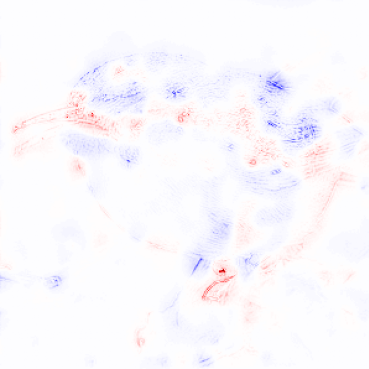}} &  
    \frame{\includegraphics[width=0.13\textwidth]{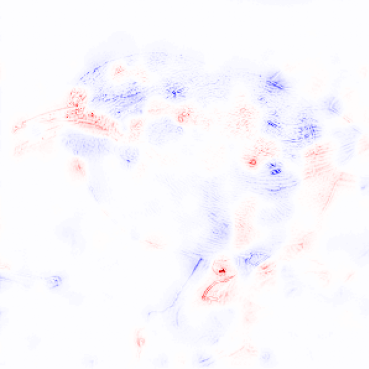}} &  
    \frame{\includegraphics[width=0.13\textwidth]{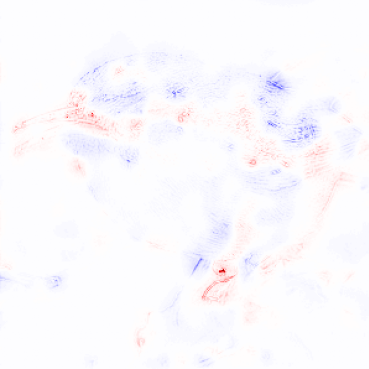}} &  
    \frame{\includegraphics[width=0.13\textwidth]{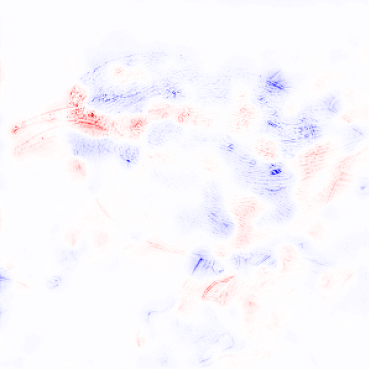}} &
    \frame{\includegraphics[width=0.13\textwidth]{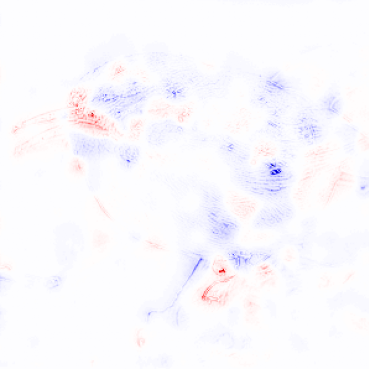}} 
 \\
 &
 &
 0.9973 &  
 0.9980 &  
 0.9928 &  
 0.9978 &  
 0.9932  
 \\
 &
    \rotatebox[origin=b]{90}{\parbox{0.05\textwidth}{Joint \mbox{without}\\sigmoid}} &
    \frame{\includegraphics[width=0.13\textwidth]{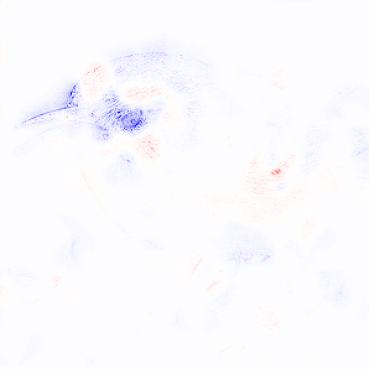}} &  
    \frame{\includegraphics[width=0.13\textwidth]{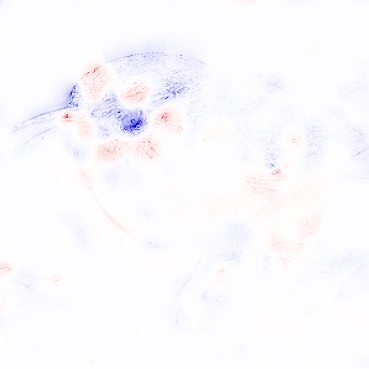}} &  
    \frame{\includegraphics[width=0.13\textwidth]{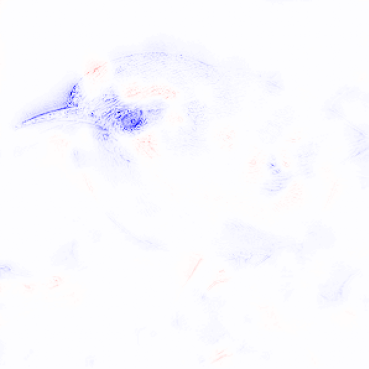}} &  
    \frame{\includegraphics[width=0.13\textwidth]{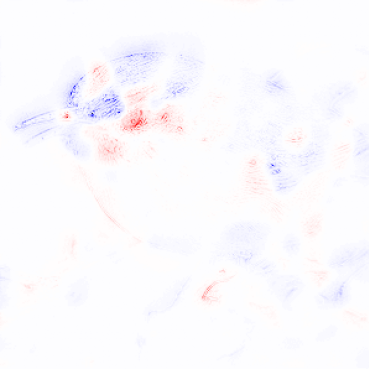}}&  
    \frame{\includegraphics[width=0.13\textwidth]{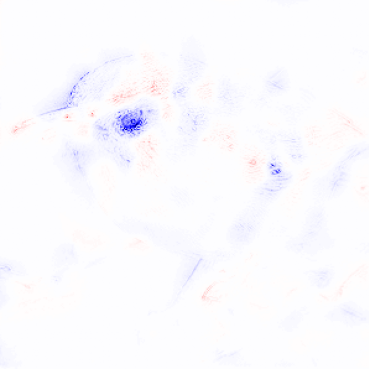}}  
 \\
 &
 &
 0.9996 &  
 0.8271 &  
 0.9918 &  
 0.9691 &  
 0.9975  
 \\
 &  
    \rotatebox[origin=b]{90}{\parbox{0.05\textwidth}{Joint with\\sigmoid}} &  
    \frame{\includegraphics[width=0.13\textwidth]{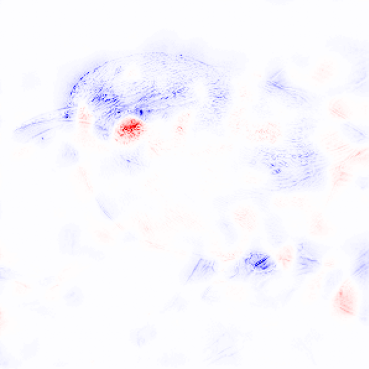}} &  
    \frame{\includegraphics[width=0.13\textwidth]{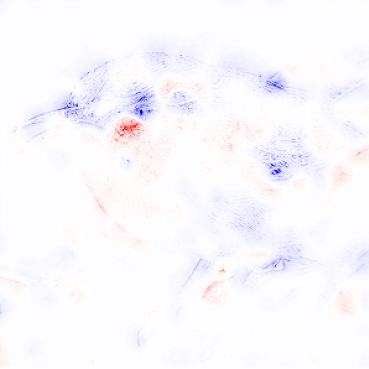}} &  
    \frame{\includegraphics[width=0.13\textwidth]{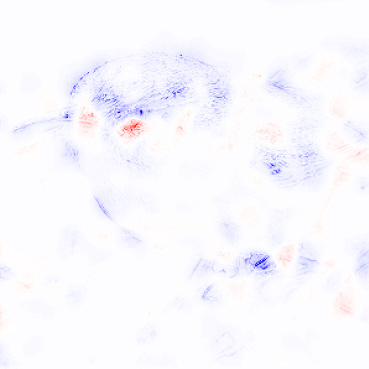}} &  
    \frame{\includegraphics[width=0.13\textwidth]{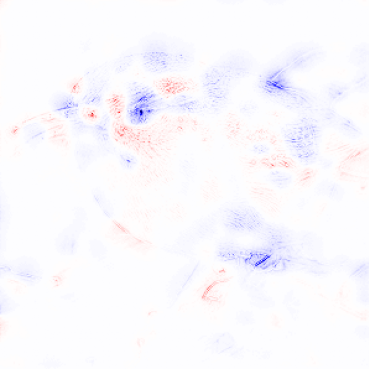}} & 
    \frame{\includegraphics[width=0.13\textwidth]{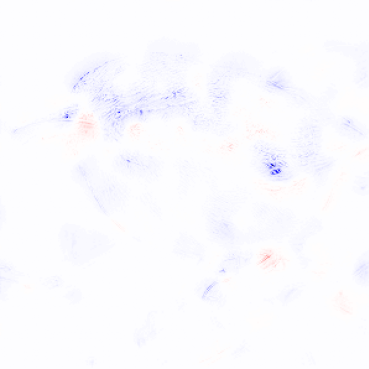}} 
 \\
 &
 &
 0.8285 &  
 0.5296 &  
 0.9890 &  
 0.9495 &  
 0.6278  
\end{tabular}
\end{center}

\caption{Concept saliency maps for the input of a Bewick Wren image and correctly predicted concepts. Positive relevance is shown in red, negative relevance is shown in blue and the predicted concept value to four decimal places, and sigmoid applied, below each saliency map. In general, relevance does not map to input features that a human would associate each concept with.}
\label{fig:CBM models concept saliency}
\end{figure*}

\begin{figure} [!ht]
    \begin{center}
    \includegraphics[width=8cm]{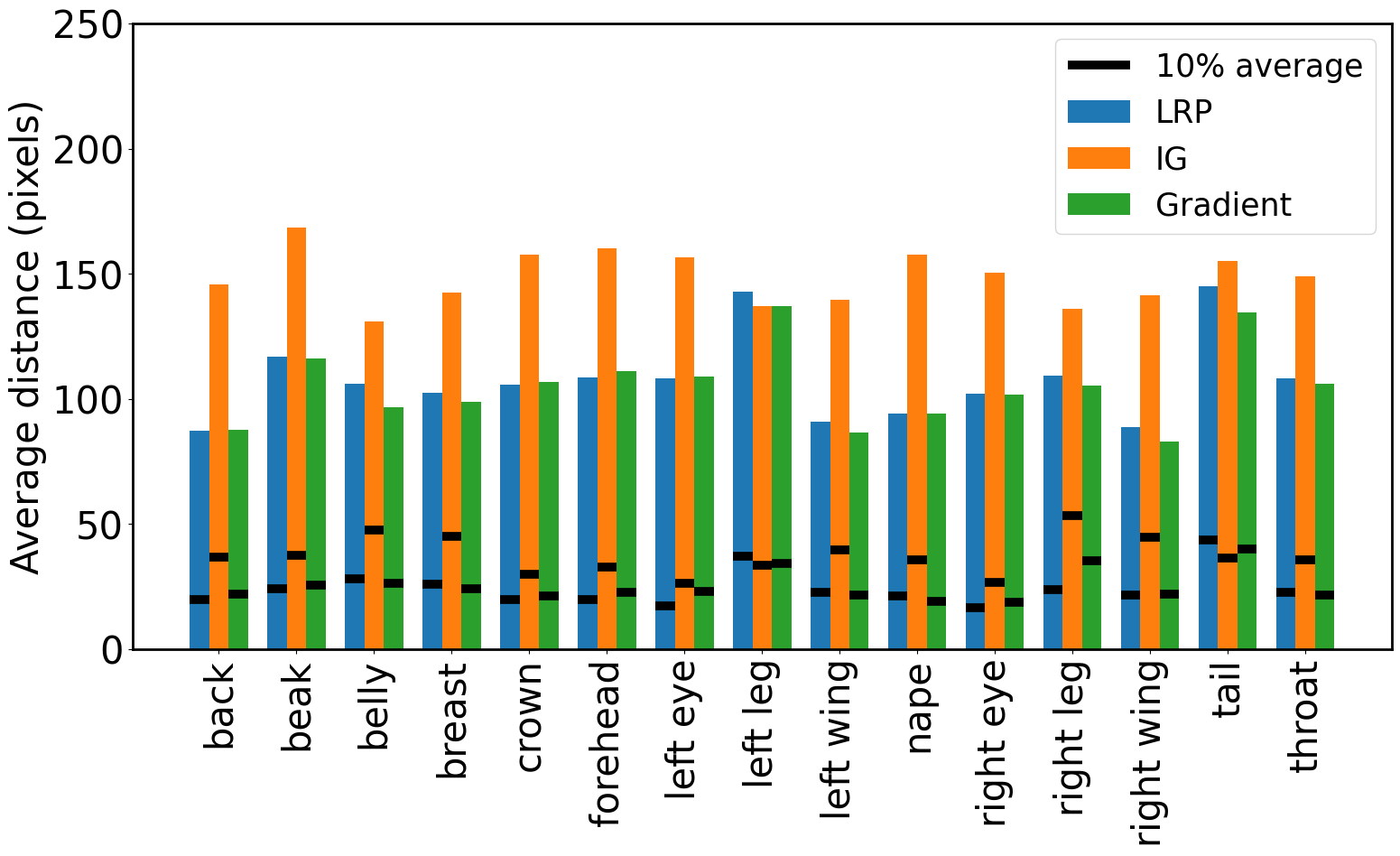}
    \end{center}
    \caption{Distance pointing game results comparing LRP, IG and a baseline gradient method. LRP and gradient has a shorter average distance for most bird parts compared to IG. This remains the same for when averaging the shortest 10\% of distances.}
    \label{fig:pointing game}
\end{figure}

Our models were trained with the same modifications of the CUB-200 2011 (CUB) dataset \citep{WahCUB_200_2011} as \citep{pmlr-v119-koh20a} which used class-level concepts with 11,788 bird images covering 200 classifications and 112 concepts.

We trained four models, one for each of the independent and sequential methods and two for the joint method (with and without a sigmoid activation between the two model parts). For the $x \rightarrow c$ model part, we used a VGG-16 architecture with batch normalisation \citep{Simonyan15}, pre-trained on ImageNet. The $c \rightarrow y$ model part is a single fully connected layer. Model performance on the test dataset is shown in Table~\ref{model summary table}.

\begin{table}
\begin{center}
    \begin{tabular}{|m{0.15\textwidth}|m{0.1\textwidth}|m{0.1\textwidth}|}
    \hline
    Training method & Classification accuracy & Concept \newline accuracy \\
    \hline\hline
    Independent & 77.51\% & 96.85\% \\
    Sequential & 75.35\% & 96.85\% \\
    Joint-without-sigmoid & 78.75\% & 96.12\% \\
    Joint-with-sigmoid & 75.35\% & 94.87\% \\
    \hline
    \end{tabular}
\end{center}
\caption{Models final classification top-1 accuracy and concept binary accuracy}
\label{model summary table}
\end{table}

%Independent model seed: 1%
%Sequential model seed: 1%
%Joint model seed: 2%
%Joint with sigmoid model seed: 1%

%Model training repository available at https://github.com/JackFurby/VGG-Concept-Bottleneck%

%After training, we converted the concept output from the $x \rightarrow c$ model part from a PyTorch ModuleList to a single linear layer. The conversion was performed post-training and does not affect model performance. Converting the concept output was not necessary for saliency map generation, but several LRP libraries \citep{NEURIPS2019_9015_captum, torchlrp} would need a modification to index the ModuleList layer before applying LRP.

\subsection{LRP Configuration} \label{lrp_config_section}

For the $x \rightarrow c$ model we use similar LRP rules to \citep{Montavon2019}. These rules are: LRP-$\alpha \beta$, where $\alpha=1$ and $\beta=0$, for the first seven convolutional layers of the model from the input, LRP-$\epsilon$ for the next six convolutional layers and LRP-0 for the top three linear layers. LRP-0 is used for the $c \rightarrow y$ model.

We used a method detailed by \citep{9190215}, changing modality for concepts, to calculate the proportion of relevance for each concept, and thus the percentage each concept contributed towards the final classification. This is possible because LRP conserves relevancy. As each concept is a single value we do not need to account for imbalance in concept proportions.

By calculating the contribution of concepts for the final classification, a human may be able to focus on the most influential concepts to a final classification and, if intervention is required, which concepts they may wish to intervene on.

\section{Results} \label{results_section}

Figure~\ref{fig:CBM models concept saliency} shows the relevance from the concept layer back to the input for a range of concepts which a human would expect to map to distinct regions of the input. Regardless of the training method used, the saliency maps indicate that the models have not learned a mapping of distinct regions in the input to concepts. Relevancy is generally distributed over the entire bird although an observation with our models is the eyes of the bird appears to be the most common feature to be highly positive or negatively relevant.

\begin{figure*}[ht!]

%result number 940%

\begin{center}

\begin{tabular}{m{0.2\textwidth} m{0.8\textwidth} }
    \multirow{8}{0.2\textwidth}{\includegraphics[width=0.15\textwidth]{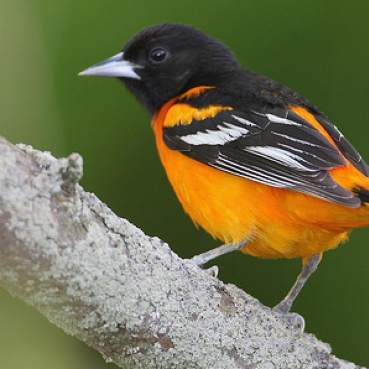}\newline \centering input} &
    \frame{\includegraphics[width=0.74\textwidth, height=0.7em]{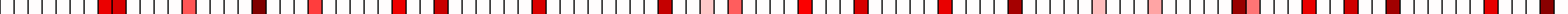}}
\\
    &
    (a) Independent
\\
    &
    \frame{\includegraphics[width=0.74\textwidth, height=0.7em]{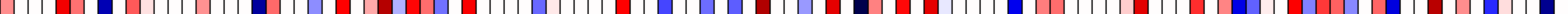}}
\\
    &
    (b) Sequential
\\
    &  
    \frame{\includegraphics[width=0.74\textwidth, height=0.7em]{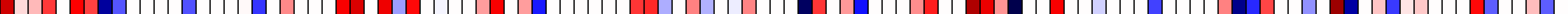}}
\\
    &
    (c) Joint-without-sigmoid
\\
    &  
    \frame{\includegraphics[width=0.74\textwidth, height=0.7em]{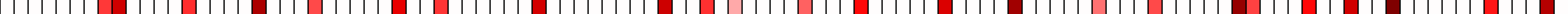}}
\\
    &
    (d) Joint-with-sigmoid
\end{tabular}

\end{center}

\caption{Final classification saliency maps for a correctly predicted Baltimore Oriole input. Each vector has 112 segments, one for each concept input. Positive relevance is shown in red and negative relevance is shown in blue. The independent and joint-with-sigmoid models only apply positive relevance to concept predicted as present. The joint-without-sigmoid and sequential models apply positive relevance to concepts predicted as not present and negative relevance to concepts predicted as present.}
\label{fig:cbm CtoY}
\end{figure*}

Concepts with similar predictions also appear to share similar saliency maps. This is evident in Figure~\ref{fig:CBM models concept saliency} with the \textit{independent and sequential} models and concepts \textit{has\_crown\_color::brown} and \textit{has\_wing\_shape::pointed-wings} which have a predicted concept value of 0.9973 and 0.9980 respectively to four decimal places. For the \textit{joint-without-sigmoid model}, \textit{has\_back\_color::brown} has a predicted concept value of 0.9918 and \textit{has\_breast\_pattern::solid} has a predicted concept value of 0.9975. The similarity between saliency maps likely means that each model has learned the same input features, can accurately predict different concepts.

Our results confirm CBMs trained on the CUB dataset do not learn distinct regions from the input to concepts, as \citep{margeloiu2021concept} showed. This is likely due to the training data or training methods not constraining the model to do so. Like regular bottleneck models \citep{4218211}, CBMs will typically only keep the most important input features, in this case, to fit the concept vector, but leave the CBM to select which input features to use. In addition, by using class-level concepts the model learns the concept vector but not if a concept is present and visible in a given sample. \citet{pmlr-v119-koh20a} version of CUB also has incorrect concepts. For example, class \textit{Mallard} has the same concept vector for males and females despite the visual differences between them. If the dataset instead had instance-level concepts, where each sample has its own concept vector only showing concepts present, we may see concept predictions that are closer to how humans would perceive them.

We also evaluated IG with a SmoothGrad noise tunnel \citep{DBLP:journals/corr/SmilkovTKVW17} using a batch size of 25 and a standard deviation of 0.2, similar to \citep{margeloiu2021concept}, using a quantitative evaluation method in comparison with LRP and a baseline gradient method \citep{Simonyan14a}. For our evaluation we modified the pointing game \citep{pointingGame} which counts hits and misses whether the most salient point of a given saliency map was within a defined region, the ground truth, resulting in an accuracy measurement. Our version, called distance pointing game, averages the distance between the most salient point of a saliency map and the ground truth point. (This was necessary because CUB does not provide bird part bounding boxes.) Our technique does not replace the pointing game, but instead, it satisfies a different situation; when you have ground truth points. By using our evaluation technique, we can quantify whether a saliency technique for a given model's output is primarily focusing on a ground truth point. We can also rank saliency techniques or models, which enables us to analyse further.

We measured the average distance using our independent model, due to that model having the highest concept accuracy, and the validation dataset split. Results are shown in Figure~\ref{fig:pointing game}. IG has around a 3rd higher average distance compared to both LRP and the baseline gradient for most bird parts while LRP and the baseline have similar average distances. To remove noisy saliency maps we also show the average of the shortest 10\% of distances which follows the same story as the overall distance averages As LRP with our rule setup groups relevance to input regions, while IG applies relevance on a pixel-by-pixel basis \cite{9369420}, LRP saliency maps are filtering out noisy relevance from the input image. However, the average distance hovers around 100 pixels away from the ground truth point with LRP and, considering the input images are 299 by 299 pixels in size, this could still fall outside of the concept in the input image, adding to what we observed in Figure~\ref{fig:CBM models concept saliency} with relevance generally covering the entire bird.

Applying LRP to the $c \rightarrow y$ model, unlike the $x \rightarrow c$ model, we can produce saliency maps with closer alignment to human decision making. Figure~\ref{fig:cbm CtoY} presents LRP saliency maps for the $c \rightarrow y$ models where they show the independent and joint-with-sigmoid training methods have learned a mapping from predicted concepts to classification that exclusively uses concepts predicted as present for the final classification. Both the sequential model and joint-without-sigmoid model applies relevance to concepts predicted as present and not present but with concept predicted as not present having positive relevance, while concepts predicted as present have negative relevance. These models appear to have learned a mapping from the concept vector prioritising the absence of concepts rather than the presence of them. Relevance is not flipped for all samples in the test dataset for these two models although it occurs often enough for it to be noteworthy. Saying for certain why the model appears to apply positive relevance to concepts that are not present and, if this is the general case of CBMs, or just models using the CUB dataset, remains an open question.

As discussed earlier, LRP enables us to calculate the contribution of each predicted concept to the final classification. For the same input as used in Figure~\ref{fig:cbm CtoY}, the top three concepts contributing to the final class predicted with the independent model are as follows: \textit{has\_upperparts\_color::white} at 6.04\%, \textit{has\_primary\_color::yellow} at 5.83\%, and \textit{has\_tail\_pattern::multi-colored} at 5.39\% with a total of 38 concepts contributing to the final classification. By calculating the concept contributions we are revealing the $c \rightarrow y$ model part reasoning towards the final classification such that a human can take this into their own decision-making when interacting with a CBM.

Additional $x \rightarrow c$ and $c \rightarrow y$ results have been included in the appendix.

\section{Conclusion and Future Work}  \label{conclusion_section}

This paper evaluates CBMs using the LRP explanation technique. LRP explanations reveal that concepts do not map to distinct regions in the input space, similar to previous work with IG explanations. However, relevance from the final classification back to the concept vector shows the model has mapped these as expected for some CBM training methods. (Exceptionally, the sequential training and joint-without-sigmoid methods applies positive relevance to concepts not predicted as present and negative relevance to ones predicted as present.) We demonstrate the ability to calculate proportional concept contribution to final classifications. Both this and the saliency maps generated from the final classification to the concept vector should aid a human user to focus on the most important concepts and improve their mental model of the CBM error boundaries.

Future work will focus on instance-level concepts to analyse if the dataset alone can confine the model to learn distinct features in the input space or whether a new training method is required. A suitable dataset will require well-labelled concepts. With the challenge of accurate concept labelling, as we have seen with CUB, synthetic datasets may be a viable option. Selecting a dataset without a 1 to 1 correlation between the concepts and final classification, such as CelebA \citep{liu2015faceattributes}, has also yet to be explored for relevancy visualisations. In addition, a human study should be conducted to analyse the effectiveness and quality of the saliency maps and concept proportion contribution when used with CBMs.

Separate to saliency maps, it would be beneficial to remove concept(s) from an input to measure changes in concept and final classification predictions. This may be measured using techniques such as Remove and Retrain \citep{10.5555/3454287.3455160} to avoid out of distribution samples affecting results.

\section*{Acknowledgements}

We thank Liam Hiley for the support he provided in the early stages of this research. In particular, for guidance on adding LRP capabilities to the models.

This research is funded by the UK Engineering and Physical Sciences Research Council (EPSRC) and IBM UK via an Industrial CASE (ICASE) award.

% References and End of Paper
% These lines must be placed at the end of your paper
\bibliography{furby}

\newpage

\appendix

\section{Appendix}

\section{Models}

We trained a total of four models for use in the paper. Each of these had the same architecture of a VGG-16 Deep Neural Network for the $x \rightarrow c$ model part and a single linear layer for the $c \rightarrow y$ model part. Each model has been trained on the same modified CUB dataset a total of three times with the highest-performing models being used to generate results. After training, we converted the $c \rightarrow y$ model part from a PyTorch ModuleList to a single linear layer. The conversion was performed post-training and does not affect model performance. Converting the concept output was not necessary for saliency map generation, but several Layer-wise Relevancy Propagation (LPR) \citep{10.1371/journal.pone.0130140} libraries \citep{NEURIPS2019_9015_captum, torchlrp} would need a modification to index the ModuleList layer before applying LRP without the conversion.

\section{Input to concept vector saliency maps}

To investigate input to concept vector relevancy we have generated saliency maps from each concept output backwards through the $x \rightarrow c$ model part to the input image. All figures use the explanation technique LRP which uses a composite set of rules (See subsection \ref{lrp_config_section}). In this section, we have repeated Figure~{\ref{fig:CBM models concept saliency}} from the paper, but as a full-page version, and included three additional examples. Each figure includes 5 saliency maps for each of the three $x \rightarrow c$ model parts and concept presence predictions where a prediction is between 0 and 1 with 0 meaning the model is confident a given concept is not present and 1 meaning the model is confident a concept is present. A value below 0.5 is counted as not present and of 0.5 or higher as present.

We show several concept predictions that cover concepts correctly predicted as present, concepts incorrectly predicted as present, concepts correctly predicted as not present and concepts incorrectly predicted as not present. Concepts correctly predicted as present are all concepts in Figure~{\ref{fig:CBM models concept saliency}}, \textit{has\_bill\allowbreak\_shape::dagger} and \textit{has\_underparts\_color::white} in Figure~{\ref{fig:second CBM models concept saliency}} and \textit{has\_bill\_length::shorter\allowbreak\_than\_head} in Figure~{\ref{fig:third CBM models concept saliency}}. Concept incorrectly predicted as present are \textit{has\_tail\_pattern::solid} in Figure~{\ref{fig:second CBM models concept saliency}} and \textit{has\_upperparts\_color::brown}, \textit{has\_breast\allowbreak\_pattern::striped}, \textit{has\_bill\_color\allowbreak::grey} and \textit{has\_breast\_pattern::striped} in Figure~{\ref{fig:third CBM models concept saliency}}. Concepts correctly predicted as not present include \textit{has\_wing\_pattern::spotted} in Figure~{\ref{fig:second CBM models concept saliency}} and \textit{has\_bill\_shape::dagger} and \textit{has\_wing\_color::grey} in Figure~{\ref{fig:fourth models concept saliency}}. Finally, concepts incorrectly predicted as not present are \textit{has\_wing\_pattern::multi-colored} in Figure~{\ref{fig:second CBM models concept saliency}} and \textit{has\_leg\_color\allowbreak::buff}, \textit{has\_underparts\allowbreak\_color::white} and \textit{has\_forehead\_color::black} in Figure~{\ref{fig:fourth models concept saliency}}.

The general case for saliency maps from the concept vector back to the input image is the salient regions, input features the model used for prediction(s), do not align to what a human may attribute relevance to, e.g. a human may base the presence of a concept for a bird leg on the appearance of the bird leg they can see. Instead, we can see the model makes concept predictions from the entire bird such as in Figure~{\ref{fig:CBM models concept saliency}} for the independent and sequential models, or seemingly unrelated parts of the bird image as shown in Figure~{\ref{fig:second CBM models concept saliency}} where the eye of the bird is a particularly important feature to the concept predictions. With our models and LRP explanations, the eye of the bird appears to be a common input feature to receive relevance.

Concepts with similar prediction values often share similar saliency maps, as shown in Figure~{\ref{fig:CBM models concept saliency}} for the Independent and Sequential models and the concepts has\_upperparts\_color\allowbreak::brown, has\_breast\_pattern::striped and has\_bill\_length::shorter\_than\_head for the joint-with-sigmoid model in Figure~{\ref{fig:third CBM models concept saliency}}. Although not a perfect match, and not occurring every time, positive and negative relevance can reverse from concept to concept for the same input image and model, such that if a concept is predicted as present, then the areas that are positively relevant may be negatively relevant for a concept predicted as not present. An example of this can be seen in Figure~{\ref{fig:second CBM models concept saliency}} for the model joint-without-sigmoid and the concepts has\_underparts\_color::white and has\_wing\_pattern::spotted.

Rarely concepts can appear to map to regions aligned with a human's own understanding of a concept such as in Figure~{\ref{fig:second CBM models concept saliency}} for the concept has\_bill\_shape\allowbreak::dagger and the model joint-with-sigmoid. In isolation, this may misguide a human user to believe the model has made a prediction using the correct input features when in reality the same input features are also being used for other concept predictions which can be seen in some of the other saliency maps for the same input image and model.

Finally, from these saliency maps, we can see the models do not need to see the presence of a bird part to make a prediction about it, such as with the concept has\_bill\_length::shorter\_than\allowbreak\_head in Figure~{\ref{fig:third CBM models concept saliency}}.

\begin{landscape}

\begin{figure}[ht!]
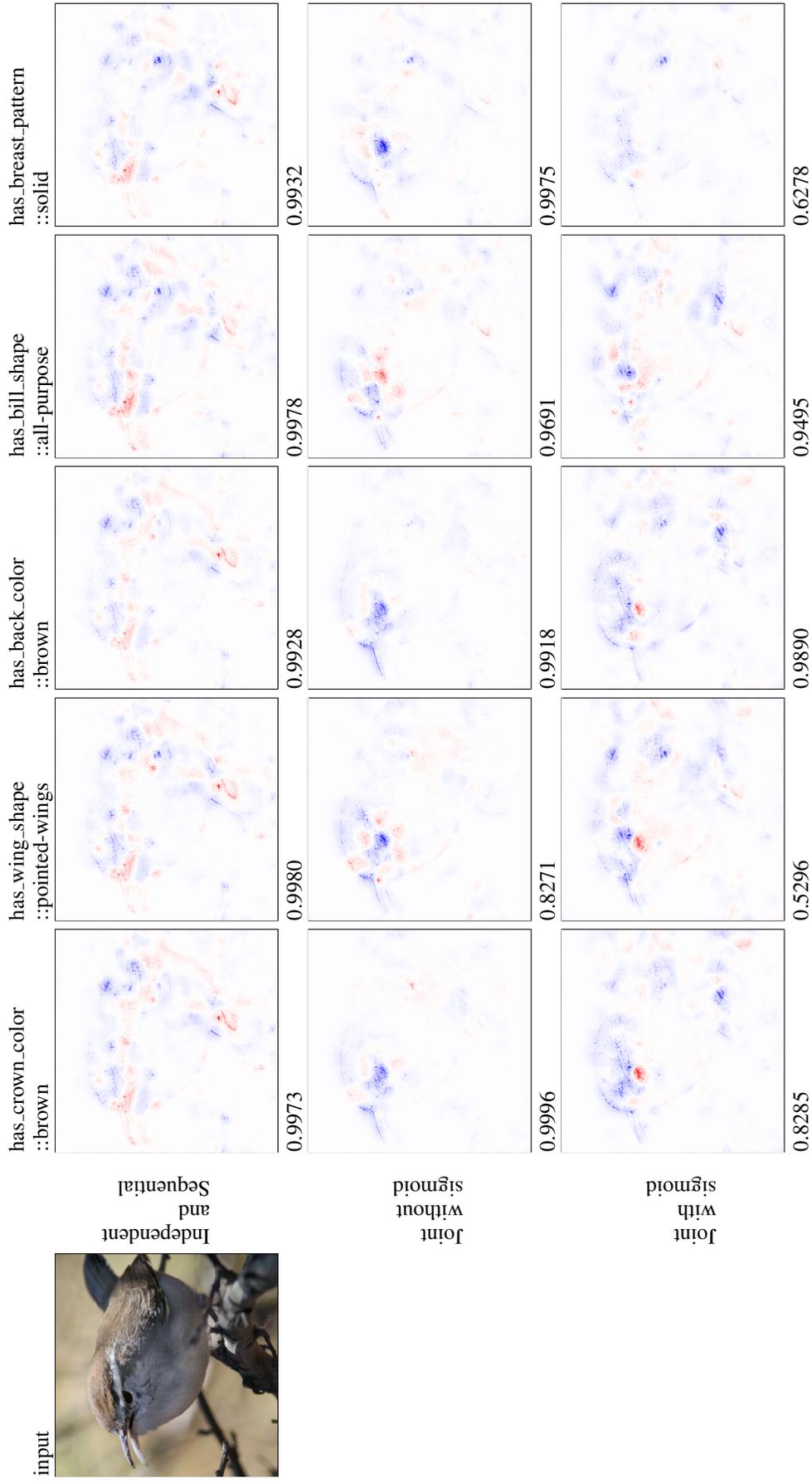


%sample ID: 1929%

\begin{center}
\begin{tabular}{m{0.185\textwidth} b{0.06\textwidth} m{0.185\textwidth} m{0.185\textwidth} m{0.185\textwidth} m{0.185\textwidth} m{0.185\textwidth} m{0.185\textwidth} }
    \vspace{4mm}input &
    &
    has\_crown\_color\newline::brown &  
    has\_wing\_shape\newline::pointed-wings &  
    has\_back\_color\newline::brown &  
    has\_bill\_shape\newline::all-purpose &  
    has\_breast\_pattern\newline::solid &
 \\
    \frame{\includegraphics[width=0.2\textwidth]{media/XtoC/bewick_wren/input.png}} &  
    \rotatebox[origin=b]{90}{\parbox{0.2\linewidth}{Independent and\\Sequential}} &  
    \frame{\includegraphics[width=0.2\textwidth]{media/XtoC/bewick_wren/independent-103.png}} &  
    \frame{\includegraphics[width=0.2\textwidth]{media/XtoC/bewick_wren/independent-77.png}} &  
    \frame{\includegraphics[width=0.2\textwidth]{media/XtoC/bewick_wren/independent-25.png}} &  
    \frame{\includegraphics[width=0.2\textwidth]{media/XtoC/bewick_wren/independent-2.png}} &
    \frame{\includegraphics[width=0.2\textwidth]{media/XtoC/bewick_wren/independent-22.png}} &
 \\
 &
 &
 0.9973 &  
 0.9980 &  
 0.9928 &  
 0.9978 &  
 0.9932  
 \\
 &
    \rotatebox[origin=b]{90}{\parbox{0.2\linewidth}{Joint \mbox{without}\\sigmoid}} &
    \frame{\includegraphics[width=0.2\textwidth]{media/XtoC/bewick_wren/joint-103.png}} &  
    \frame{\includegraphics[width=0.2\textwidth]{media/XtoC/bewick_wren/joint-77.png}} &  
    \frame{\includegraphics[width=0.2\textwidth]{media/XtoC/bewick_wren/joint-25.png}} &  
    \frame{\includegraphics[width=0.2\textwidth]{media/XtoC/bewick_wren/joint-2.png}}&  
    \frame{\includegraphics[width=0.2\textwidth]{media/XtoC/bewick_wren/joint-22.png}} &  
 \\
 &
 &
 0.9996 &  
 0.8271 &  
 0.9918 &  
 0.9691 &  
 0.9975  
 \\
 &  
    \rotatebox[origin=b]{90}{\parbox{0.2\linewidth}{Joint with\\sigmoid}} &  
    \frame{\includegraphics[width=0.2\textwidth]{media/XtoC/bewick_wren/jointSig-103.png}} &  
    \frame{\includegraphics[width=0.2\textwidth]{media/XtoC/bewick_wren/jointSig-77.png}} &  
    \frame{\includegraphics[width=0.2\textwidth]{media/XtoC/bewick_wren/jointSig-25.png}} &  
    \frame{\includegraphics[width=0.2\textwidth]{media/XtoC/bewick_wren/jointSig-2.png}} & 
    \frame{\includegraphics[width=0.2\textwidth]{media/XtoC/bewick_wren/jointSig-22.png}} &
\\
 &
 &
 0.8285 &  
 0.5296 &  
 0.9890 &  
 0.9495 &  
 0.6278  
\end{tabular}
\end{center}

\caption{larger version of Figure~{\ref{fig:CBM models concept saliency}} in the paper. All concepts were correctly predicted as present. The input class was Bewick wren.}
\label{fig:first CBM models concept saliency}
\end{figure}

\end{landscape}

\begin{landscape}

\begin{figure}[ht!]

%sample ID: 1447%

\begin{center}
\begin{tabular}{m{0.185\textwidth} b{0.06\textwidth} m{0.185\textwidth} m{0.185\textwidth} m{0.185\textwidth} m{0.185\textwidth} m{0.185\textwidth} m{0.185\textwidth} }
    \vspace{4mm}input &
    &
    has\_bill\_shape\newline::dagger &  
    has\_underparts\newline\_color\newline::white &  
    has\_wing\_pattern\newline::spotted &  
    has\_tail\_pattern\newline::solid &  
    has\_wing\_pattern\newline::multi-colored &
 \\
    \frame{\includegraphics[width=0.2\textwidth]{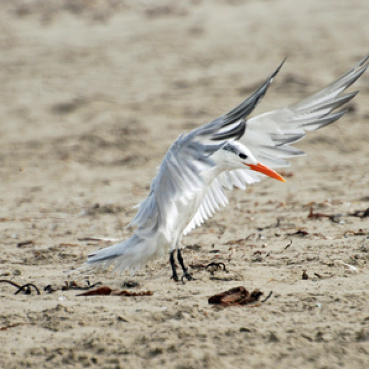}} &  
    \rotatebox[origin=b]{90}{\parbox{0.2\linewidth}{Independent and\\Sequential}} &  
    \frame{\includegraphics[width=0.2\textwidth]{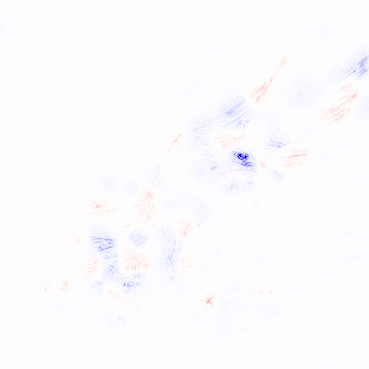}} &  
    \frame{\includegraphics[width=0.2\textwidth]{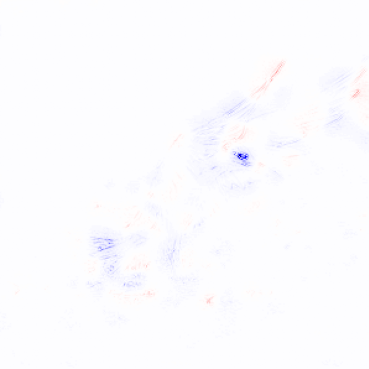}} &  
    \frame{\includegraphics[width=0.2\textwidth]{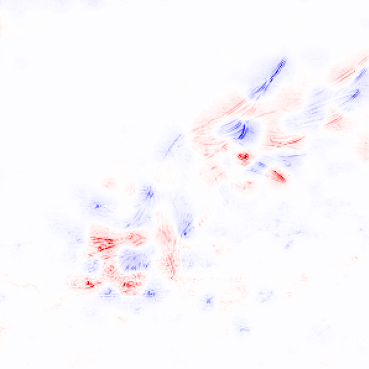}} &  
    \frame{\includegraphics[width=0.2\textwidth]{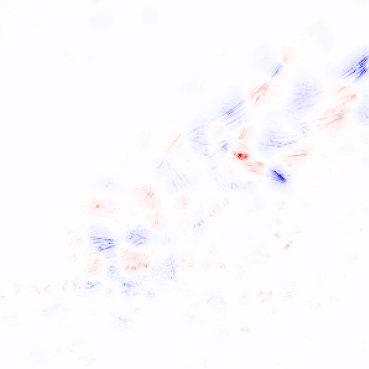}} &
    \frame{\includegraphics[width=0.2\textwidth]{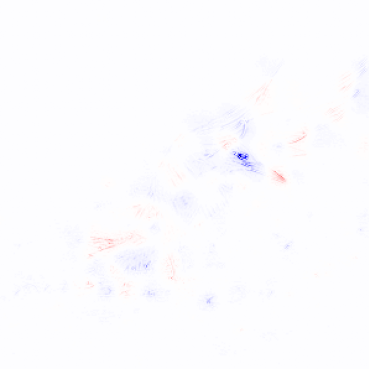}} &
 \\
 &
 &
 0.9894 &  
 0.9742 &  
 0.0007 &  
 0.6506 &  
 0.0830  
 \\
 &
    \rotatebox[origin=b]{90}{\parbox{0.2\linewidth}{Joint \mbox{without}\\sigmoid}} &
    \frame{\includegraphics[width=0.2\textwidth]{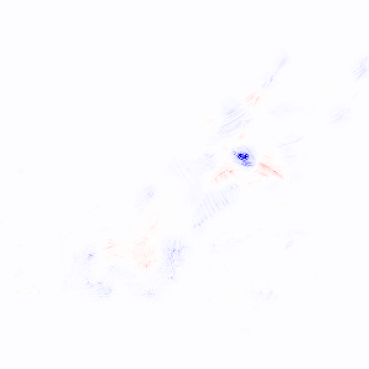}} &  
    \frame{\includegraphics[width=0.2\textwidth]{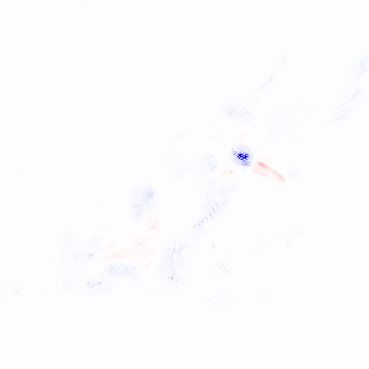}} &  
    \frame{\includegraphics[width=0.2\textwidth]{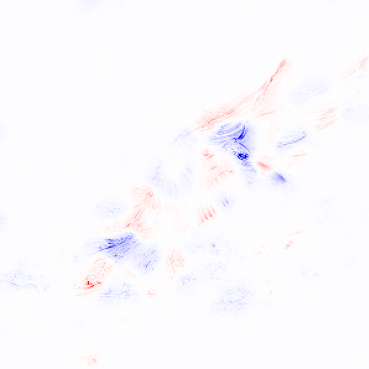}} &  
    \frame{\includegraphics[width=0.2\textwidth]{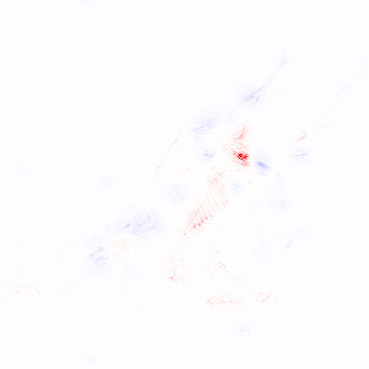}}&  
    \frame{\includegraphics[width=0.2\textwidth]{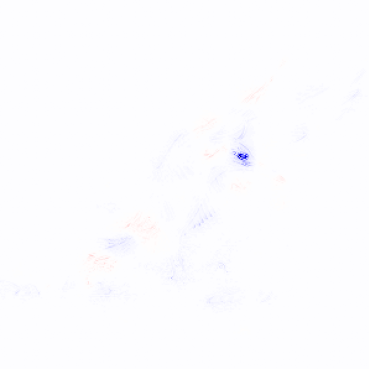}} &  
\\
 &
 &
 0.9993 &  
 1.0000 &  
 0.0002 &  
 0.8989 &  
 0.0257  
 \\
 &  
    \rotatebox[origin=b]{90}{\parbox{0.2\linewidth}{Joint with\\sigmoid}} &  
    \frame{\includegraphics[width=0.2\textwidth]{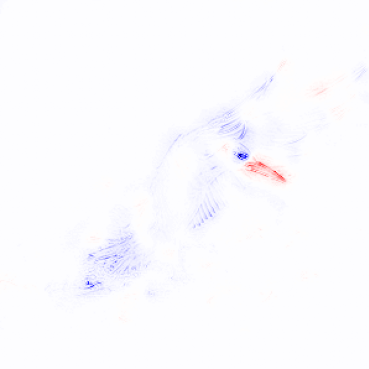}} &  
    \frame{\includegraphics[width=0.2\textwidth]{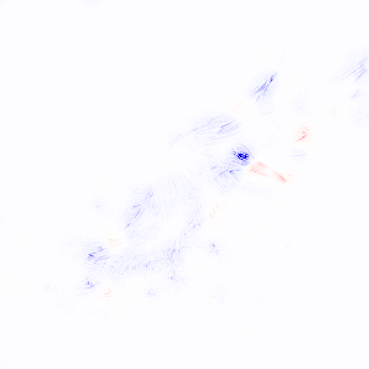}} &  
    \frame{\includegraphics[width=0.2\textwidth]{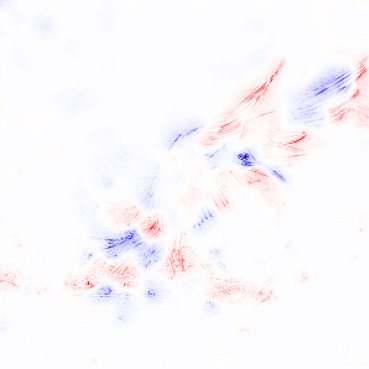}} &  
    \frame{\includegraphics[width=0.2\textwidth]{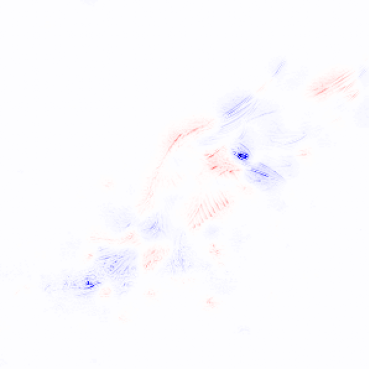}} & 
    \frame{\includegraphics[width=0.2\textwidth]{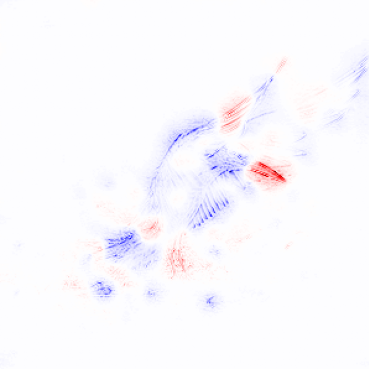}} &
\\
 &
 &
 0.9966 &  
 0.9934 &  
 0.0060 &  
 0.6646 &  
 0.0397  
\end{tabular}
\end{center}

\caption{Concepts are a mixture of present, not present, correctly predicted and not correctly predicted. The input class was Elegant\_tern.}
\label{fig:second CBM models concept saliency}
\end{figure}

\end{landscape}

\begin{landscape}

\begin{figure}[ht!]

%sample ID: 1238%

\begin{center}
\begin{tabular}{m{0.185\textwidth} b{0.06\textwidth} m{0.185\textwidth} m{0.185\textwidth} m{0.185\textwidth} m{0.185\textwidth} m{0.185\textwidth} m{0.185\textwidth} }
    \vspace{4mm}input &
    &
    has\_upperparts\_color\newline::brown &  
    has\_breast\_pattern\newline::striped &  
    has\_bill\_color\newline::grey &  
    has\_breast\_pattern\newline::striped &  
    has\_bill\_length\newline::shorter\_than\_head &
 \\
    \frame{\includegraphics[width=0.2\textwidth]{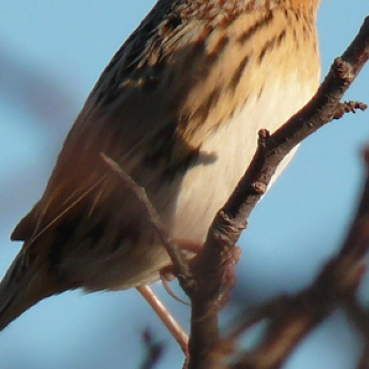}} &  
    \rotatebox[origin=b]{90}{\parbox{0.2\linewidth}{Independent and\\Sequential}} &  
    \frame{\includegraphics[width=0.2\textwidth]{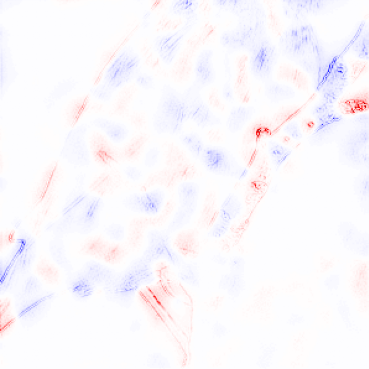}} &  
    \frame{\includegraphics[width=0.2\textwidth]{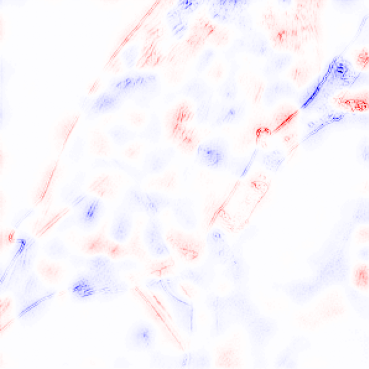}} &  
    \frame{\includegraphics[width=0.2\textwidth]{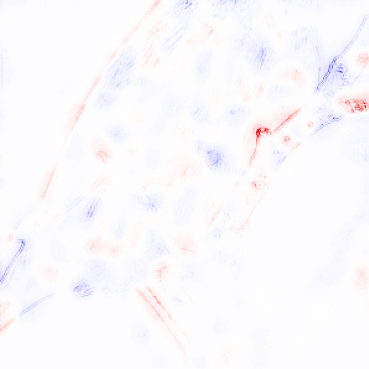}} &  
    \frame{\includegraphics[width=0.2\textwidth]{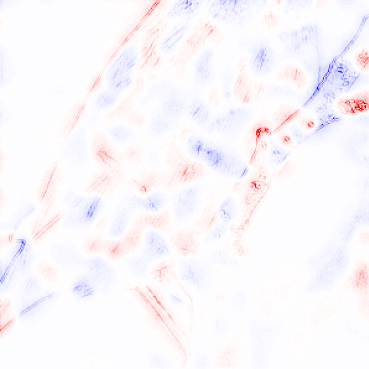}} &
    \frame{\includegraphics[width=0.2\textwidth]{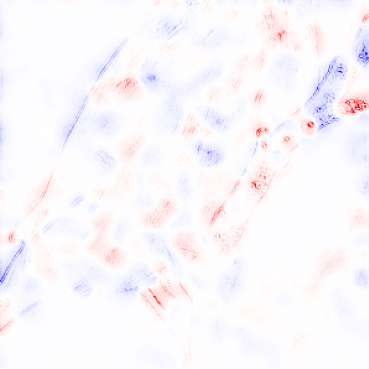}} &
 \\
 &
 &
 0.9331 &  
 0.6930 &  
 0.8301 &  
 0.7508 &  
 0.9708  
 \\
 &
    \rotatebox[origin=b]{90}{\parbox{0.2\linewidth}{Joint \mbox{without}\\sigmoid}} &
    \frame{\includegraphics[width=0.2\textwidth]{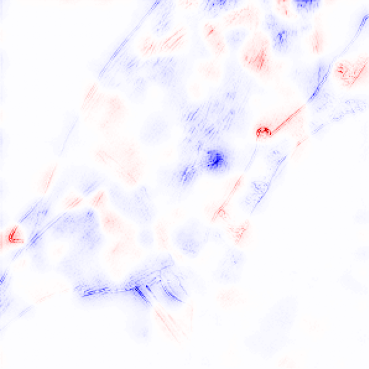}} &  
    \frame{\includegraphics[width=0.2\textwidth]{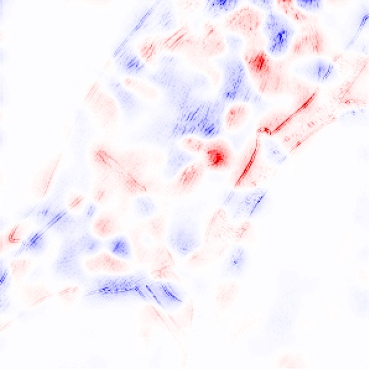}} &  
    \frame{\includegraphics[width=0.2\textwidth]{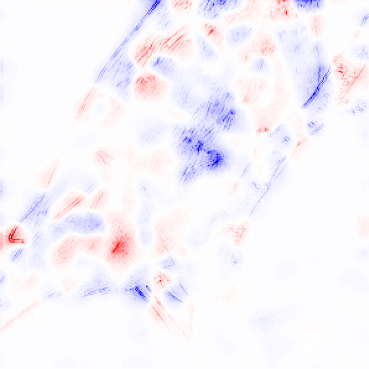}} &  
    \frame{\includegraphics[width=0.2\textwidth]{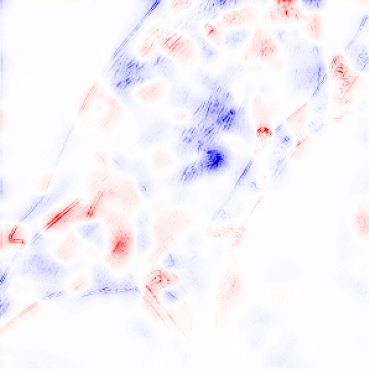}}&  
    \frame{\includegraphics[width=0.2\textwidth]{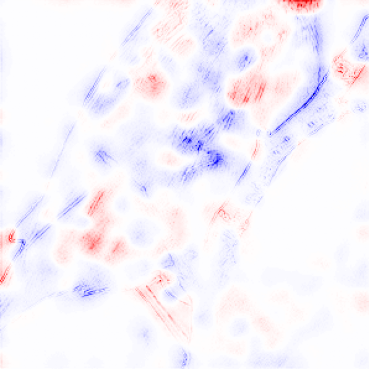}} &  
 \\
 &
 &
 0.8966 &  
 0.8097 &  
 0.7740 &  
 0.7546 &  
 0.9990  
 \\
 &  
    \rotatebox[origin=b]{90}{\parbox{0.2\linewidth}{Joint with\\sigmoid}} &  
    \frame{\includegraphics[width=0.2\textwidth]{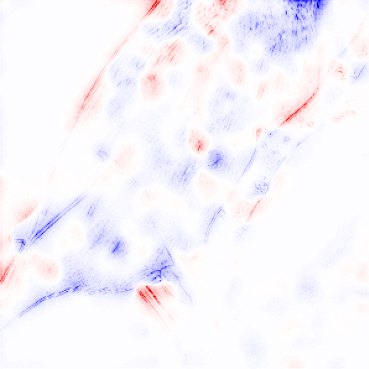}} &  
    \frame{\includegraphics[width=0.2\textwidth]{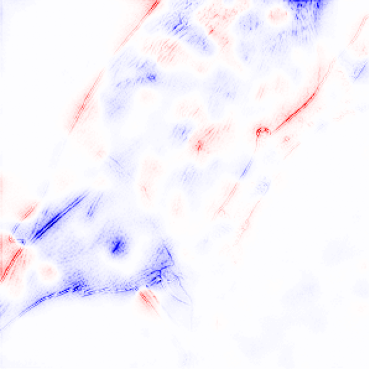}} &  
    \frame{\includegraphics[width=0.2\textwidth]{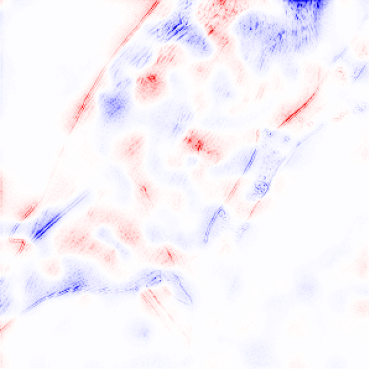}} &  
    \frame{\includegraphics[width=0.2\textwidth]{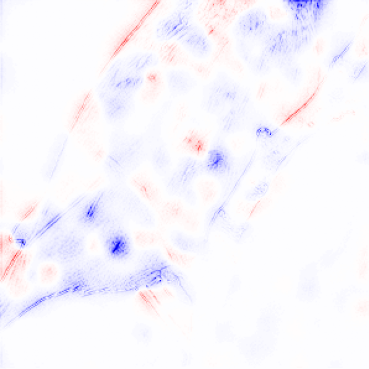}} & 
    \frame{\includegraphics[width=0.2\textwidth]{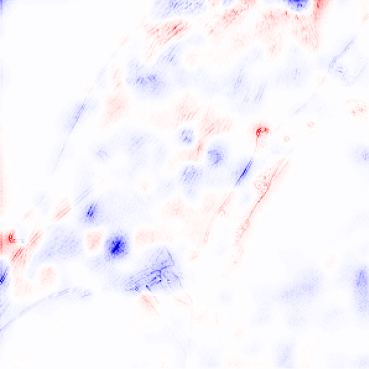}} &
\\
 &
 &
 0.9188 &  
 0.9815 &  
 0.7584 &  
 0.9484 &  
 0.9674
\end{tabular}
\end{center}

\caption{Saliency maps for $x \rightarrow c$ models with the input of a Le Conte Sparrow image. Concepts shown are either incorrectly predicted as present or correctly predicted as present but not visible in the input image. The input class was Le Conte Sparrow.}
\label{fig:third CBM models concept saliency}
\end{figure}

\end{landscape}

\begin{landscape}

\begin{figure}[ht!]

%sample ID: 1257%

\begin{center}
\begin{tabular}{m{0.185\textwidth} b{0.06\textwidth} m{0.185\textwidth} m{0.185\textwidth} m{0.185\textwidth} m{0.185\textwidth} m{0.185\textwidth} m{0.185\textwidth} }
    \vspace{4mm}input &
    &
    has\_leg\_color\newline::buff &  
    has\_underparts\_color\newline::white &  
    has\_forehead\_color\newline::black &  
    has\_bill\_shape\newline::dagger &  
    has\_wing\_color\newline::grey &
 \\
    \frame{\includegraphics[width=0.2\textwidth]{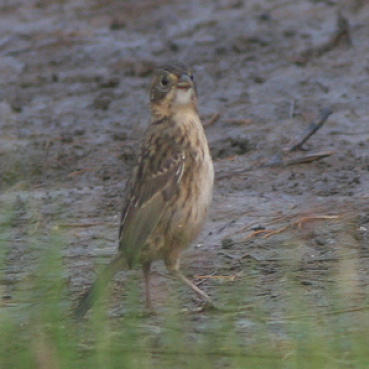}} &  
    \rotatebox[origin=b]{90}{\parbox{0.2\linewidth}{Independent and\\Sequential}} &  
    \frame{\includegraphics[width=0.2\textwidth]{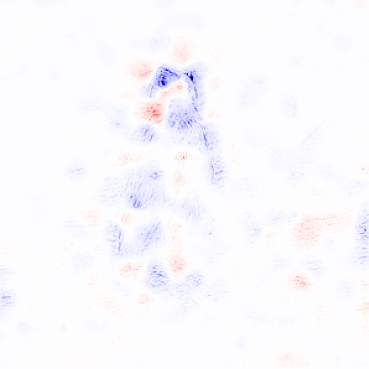}} &  
    \frame{\includegraphics[width=0.2\textwidth]{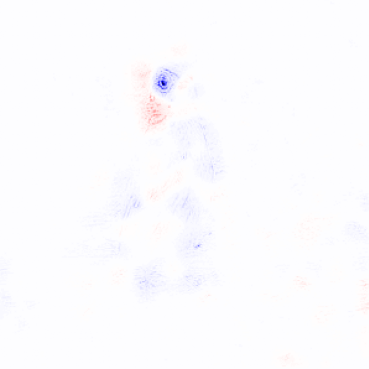}} &  
    \frame{\includegraphics[width=0.2\textwidth]{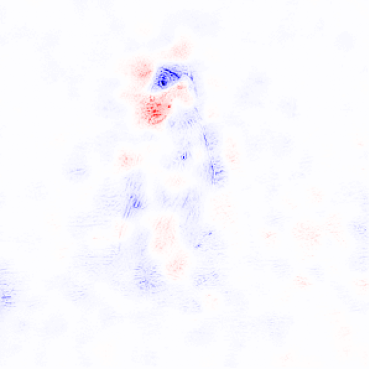}} &  
    \frame{\includegraphics[width=0.2\textwidth]{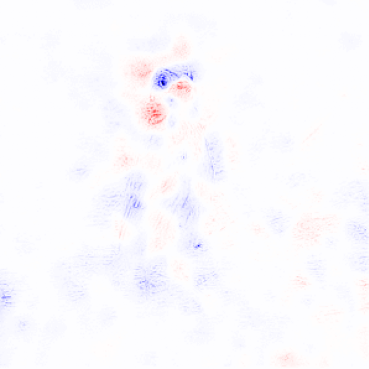}} &
    \frame{\includegraphics[width=0.2\textwidth]{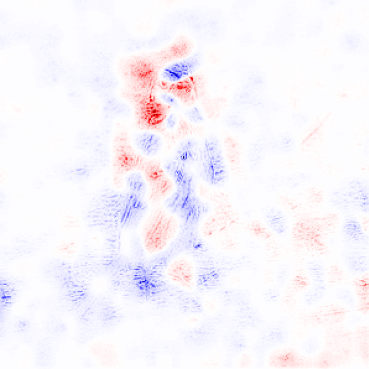}} &
 \\
 &
 &
 0.0432 &  
 0.0252 &  
 0.0073 &  
 0.0002 &  
 0.0002  
 \\
 &
    \rotatebox[origin=b]{90}{\parbox{0.2\linewidth}{Joint \mbox{without}\\sigmoid}} &
    \frame{\includegraphics[width=0.2\textwidth]{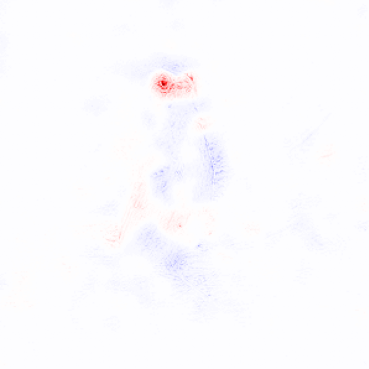}} &  
    \frame{\includegraphics[width=0.2\textwidth]{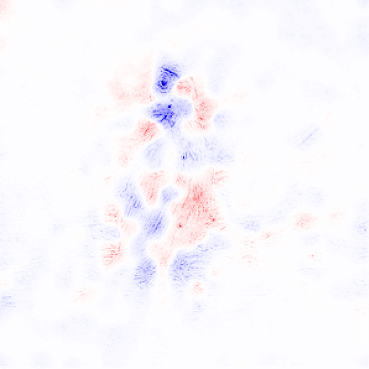}} &  
    \frame{\includegraphics[width=0.2\textwidth]{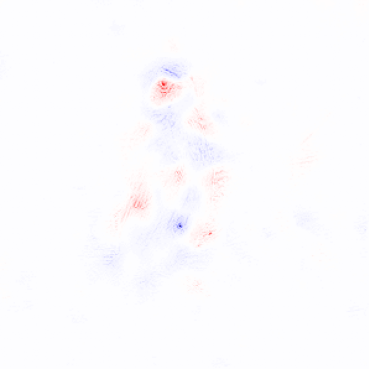}} &  
    \frame{\includegraphics[width=0.2\textwidth]{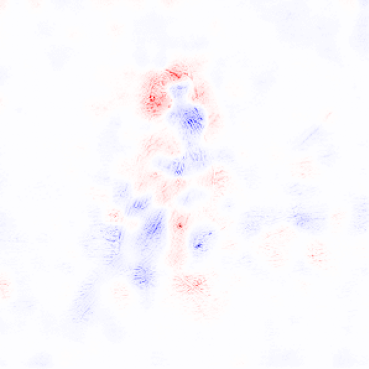}}&  
    \frame{\includegraphics[width=0.2\textwidth]{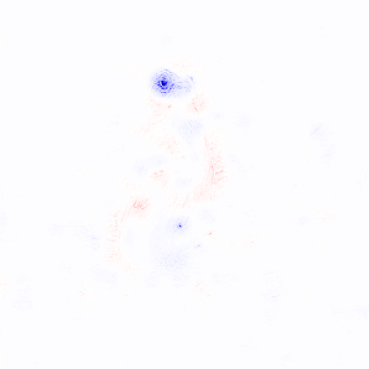}} &  
 \\
 &
 &
 0.2583 &  
 0.1353 &  
 0.3375 &  
 0.0017 &  
 0.0036  
 \\
 &    
    \rotatebox[origin=b]{90}{\parbox{0.2\linewidth}{Joint with\\sigmoid}} &  
    \frame{\includegraphics[width=0.2\textwidth]{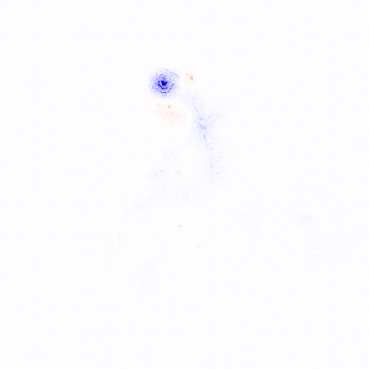}} &  
    \frame{\includegraphics[width=0.2\textwidth]{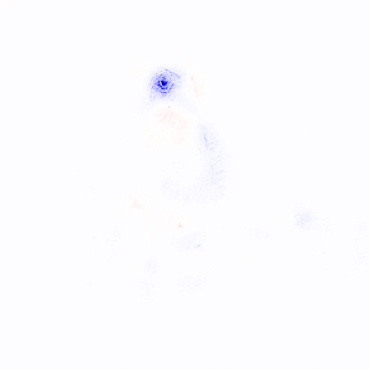}} &  
    \frame{\includegraphics[width=0.2\textwidth]{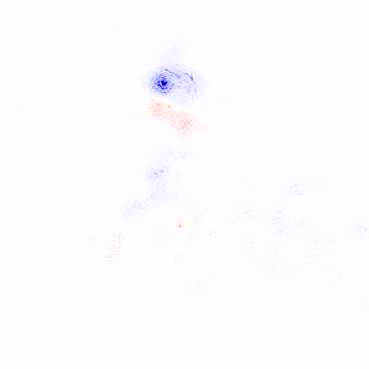}} &  
    \frame{\includegraphics[width=0.2\textwidth]{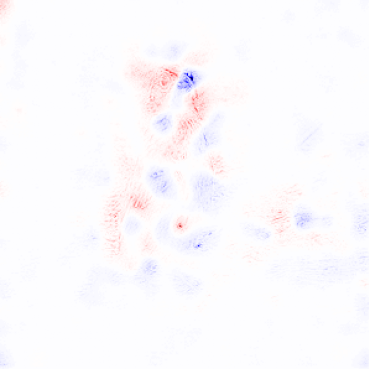}} & 
    \frame{\includegraphics[width=0.2\textwidth]{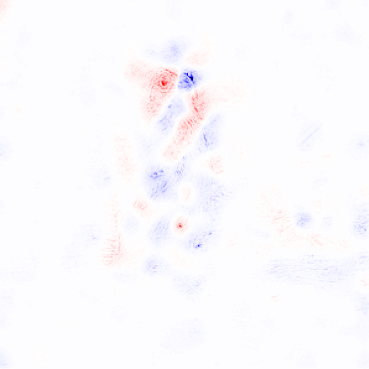}} &
\\
 &
 &
 0.4392 &  
 0.3549 &  
 0.1322 &  
 0.0044 &  
 0.0023
\end{tabular}
\end{center}

\caption{Saliency maps for $x \rightarrow c$ models with the input of a Nelson Sharp tailed Sparrow image. Concepts shown are either correctly or incorrectly predicted as not present. The input class was Nelson Sharp tailed Sparrow.}
\label{fig:fourth models concept saliency}
\end{figure}

\end{landscape}

\section{Concept vector to final classification saliency maps}

\begin{landscape}

\begin{figure}[t!]
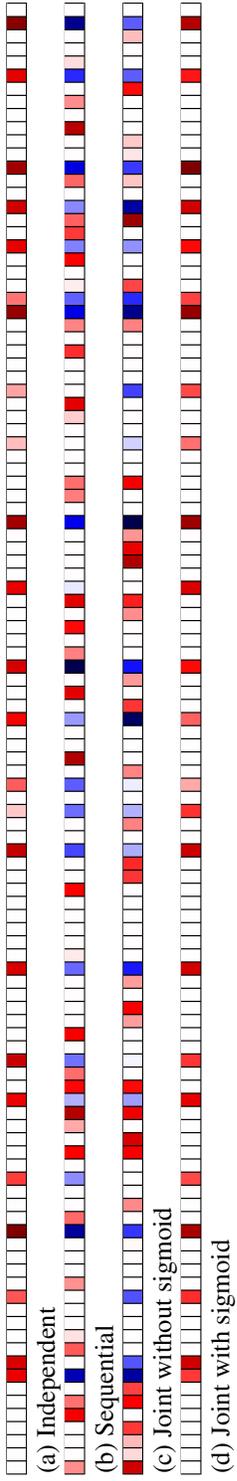


%result number 940%

\begin{center}

\begin{tabular}{m{0.2\textwidth} m{8\textwidth} }
    \multirow{8}{0.2\textwidth}{\includegraphics[width=0.2\textwidth]{media/CtoY/Baltimore_Oriole/input.png}\newline \centering input} &
    \frame{\includegraphics[width=1.1\textwidth, height=0.7em]{media/CtoY/Baltimore_Oriole/independent.png}}
\\
    &
    (a) Independent
\\
    &
    \frame{\includegraphics[width=1.1\textwidth, height=0.7em]{media/CtoY/Baltimore_Oriole/sequential.png}}
\\
    &
    (b) Sequential
\\
    &  
    \frame{\includegraphics[width=1.1\textwidth, height=0.7em]{media/CtoY/Baltimore_Oriole/joint.png}}
\\
    &
    (c) Joint without sigmoid
\\
    &  
    \frame{\includegraphics[width=1.1\textwidth, height=0.7em]{media/CtoY/Baltimore_Oriole/jointSig.png}}
\\
    &
    (d) Joint with sigmoid
\end{tabular}

\end{center}

\vspace{6mm}
\caption{larger version of Figure~{\ref{fig:cbm CtoY}} in the paper. Saliency maps for $c \rightarrow y$ for each training method where the final classification is correctly predicted as Baltimore Oriole.}
\label{fig:first cbm CtoY}
\end{figure}

\end{landscape}

\begin{landscape}

\begin{figure}[t!]

%result number 74%

\begin{center}

\begin{tabular}{m{0.2\textwidth} m{1.6\textwidth} }
    \multirow{8}{0.2\textwidth}{\includegraphics[width=0.2\textwidth]{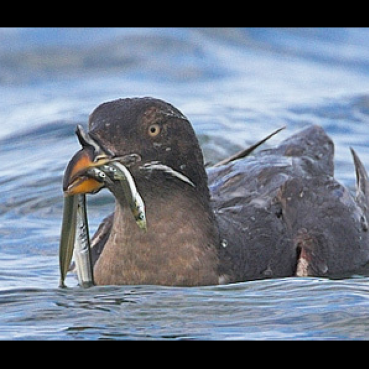}\newline \centering input} &
    \frame{\includegraphics[width=1.1\textwidth, height=0.7em]{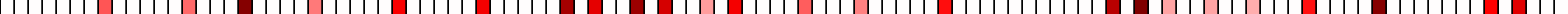}}
\\
    &
    (a) Independent
\\
    &
    \frame{\includegraphics[width=1.1\textwidth, height=0.7em]{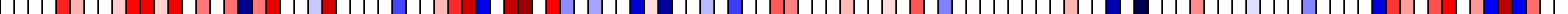}}
\\
    &
    (b) Sequential
\\
    &  
    \frame{\includegraphics[width=1.1\textwidth, height=0.7em]{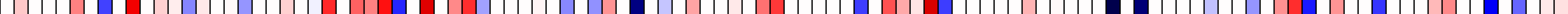}}
\\
    &
    (c) Joint without sigmoid
\\
    &  
    \frame{\includegraphics[width=1.1\textwidth, height=0.7em]{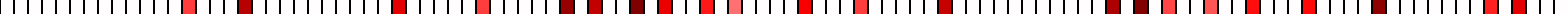}}
\\
    &
    (d) Joint with sigmoid
\end{tabular}

\end{center}

\vspace{6mm}
\caption{Saliency maps for $c \rightarrow y$ for each training method where the final classification is correctly predicted as Rhinoceros Auklet.}
\label{fig:second cbm CtoY}
\end{figure}

\end{landscape}

\begin{landscape}

\begin{figure}[t!]

%result number 610%

\begin{center}

\begin{tabular}{m{0.2\textwidth} m{1.6\textwidth} }
    \multirow{8}{0.2\textwidth}{\includegraphics[width=0.2\textwidth]{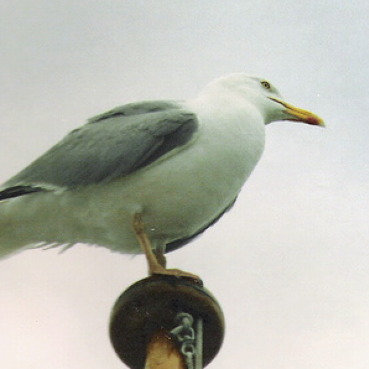}\newline \centering input} &
    \frame{\includegraphics[width=1.1\textwidth, height=0.7em]{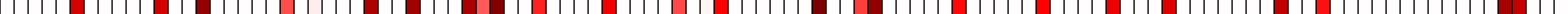}}
\\
    &
    (a) Independent
\\
    &
    \frame{\includegraphics[width=1.1\textwidth, height=0.7em]{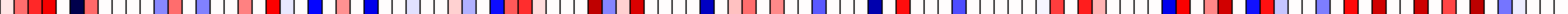}}
\\
    &
    (b) Sequential
\\
    &  
    \frame{\includegraphics[width=1.1\textwidth, height=0.7em]{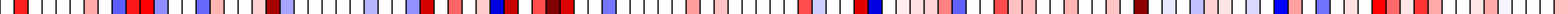}}
\\
    &
    (c) Joint without sigmoid
\\
    &  
    \frame{\includegraphics[width=1.1\textwidth, height=0.7em]{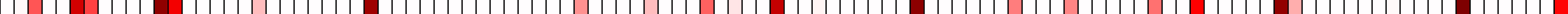}}
\\
    &
    (d) Joint with sigmoid
\end{tabular}

\end{center}

\vspace{6mm}
\caption{Saliency maps for $c \rightarrow y$ for each training method where the final classification is incorrect. The correct class is Herring gull and the predicted class was California gull for the independent model, Western gull for joint-without-sigmoid model, Nashville warbler for the joint-with-sigmoid model and Red legged kittiwake for the sequential model.}
\label{fig:third cbm CtoY}
\end{figure}

\end{landscape}

\begin{landscape}

\begin{figure}[t!]

%result number 881%

\begin{center}

\begin{tabular}{m{0.2\textwidth} m{1.6\textwidth} }
    \multirow{8}{0.2\textwidth}{\includegraphics[width=0.2\textwidth]{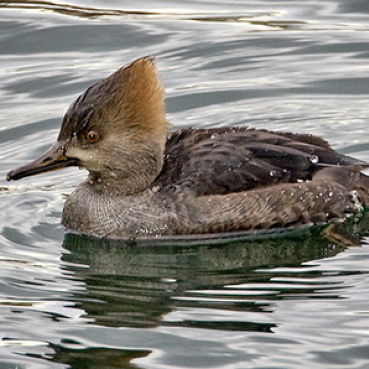}\newline \centering input} &
    \frame{\includegraphics[width=1.1\textwidth, height=0.7em]{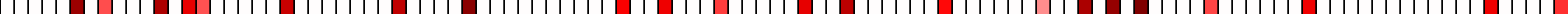}}
\\
    &
    (a) Independent
\\
    &
    \frame{\includegraphics[width=1.1\textwidth, height=0.7em]{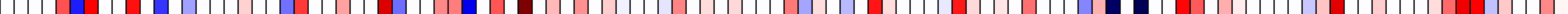}}
\\
    &
    (b) Sequential
\\
    &  
    \frame{\includegraphics[width=1.1\textwidth, height=0.7em]{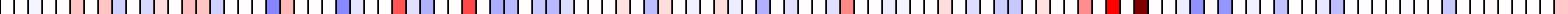}}
\\
    &
    (c) Joint without sigmoid
\\
    &  
    \frame{\includegraphics[width=1.1\textwidth, height=0.7em]{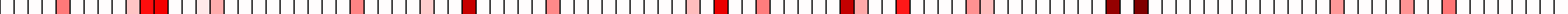}}
\\
    &
    (d) Joint with sigmoid
\end{tabular}

\end{center}

\vspace{6mm}
\caption{Saliency maps for $c \rightarrow y$ for each training method where the final classification is incorrect. The correct classification is Hooded Merganser. The predicted classifications were Red breasted Merganser for the independent model, joint-without-sigmoid model and sequential model, and Godwall for the joint-without-sigmoid model.}
\label{fig:fourth cbm CtoY}
\end{figure}

\end{landscape}

For relevance from the final classification to the concept vector, we have also employed LRP to generate saliency maps. From doing so we can see the final classification is predicted with concepts predicted as present, and in some cases, concepts predicted as not present. We have a larger version of Figure 3 from the paper and three additional examples. Each figure is accompanied by four tables with concept IDs, concept vector predictions, LRP relevance for the final classification from the concept vector, and concept contributions. The tables are sorted to display the concept with the highest contribution to the final classification first, followed by concepts in descending order of contribution. The concept vectors for each model are split into 112 segments, one for each concept in the dataset, with the segment most to the left of each vector representing the concept with ID 0 and the segment most to the right representing the concept with ID 111. Our independent model and joint-with-sigmoid model were trained with a sigmoid layer between the $x \rightarrow c$ model part and the $c \rightarrow y$ model part while the sequential model and joint-without-sigmoid model were trained without a sigmoid layer. The models correctly predicted the final classifications for the inputs in Figures~{\ref{fig:first cbm CtoY} and \ref{fig:second cbm CtoY}} but made incorrect predictions for the inputs in figures~{\ref{fig:third cbm CtoY} and \ref{fig:fourth cbm CtoY}}.

The saliency maps for the independent and joint-with-sigmoid models only attribute positive relevance to concepts predicted as present with no concept receiving negative relevance. The sequential and the joint-without-sigmoid models attribute both positive and negative relevance with this mostly resulting in concepts predicted as present receiving negative relevance. An exception to this is shown in Figure~{\ref{fig:fourth cbm CtoY}} for the joint-without-sigmoid model. Here concepts predicted as present have positive relevancy and those predicted as not present have negative relevancy or no relevance. A concept can contribute to the final classification irrespective of if the concept is positively or negatively relevant.

\onecolumn

\begin{center}
    \begin{longtable}{|m{0.08\textwidth}|m{0.38\textwidth}|m{0.1\textwidth}|m{0.12\textwidth}|m{0.12\textwidth}|}
    \hline
    Concept ID &
    Concept name &  
    Concept Value &  
    Relevancy &  
    Concept contribution \\
    \hline\hline
    18 & has\_underparts\_color::yellow & 1.0 & 0.0728 & 6.933\% \\
    110 & has\_wing\_pattern::striped & 1.0 & 0.0696 & 6.623\% \\
    88 & has\_tail\_pattern::multi-colored & 1.0 & 0.0658 & 6.27\% \\
    99 & has\_bill\_color::grey & 1.0 & 0.0628 & 5.977\% \\
    72 & has\_belly\_color::yellow & 1.0 & 0.0615 & 5.855\% \\
    47 & has\_throat\_color::black & 1.0 & 0.0525 & 4.995\% \\
    31 & has\_tail\_shape::notched\_tail & 1.0 & 0.0516 & 4.911\% \\
\end{longtable}
\end{center}

\begin{center}
    \begin{longtable}{|m{0.08\textwidth}|m{0.38\textwidth}|m{0.1\textwidth}|m{0.12\textwidth}|m{0.12\textwidth}|}
    \hline
    Concept ID &
    Concept name &  
    Concept Value &  
    Relevancy &  
    Concept contribution \\
    \hline\hline
    18 & has\_underparts\_color::yellow & 1.0 & -0.1877 & 5.833\% \\
    110 & has\_wing\_pattern::striped & 1.0 & -0.1963 & 5.464\% \\
    61 & has\_under\_tail\_color::black & 1.0 & -0.2436 & 4.412\% \\
    88 & has\_tail\_pattern::multi-colored & 1.0 & -0.1373 & 3.972\% \\
    72 & has\_belly\_color::yellow & 1.0 & -0.1321 & 3.865\% \\
    99 & has\_bill\_color::grey & 1.0 & -0.1431 & 3.845\% \\
    7 & has\_wing\_color::black & 1.0 & -0.1715 & 3.504\% \\
\end{longtable}
\end{center}

\begin{center}
    \begin{longtable}{|m{0.08\textwidth}|m{0.38\textwidth}|m{0.1\textwidth}|m{0.12\textwidth}|m{0.12\textwidth}|}
    \hline
    Concept ID &
    Concept name &  
    Concept Value &  
    Relevancy &  
    Concept contribution \\
    \hline\hline
    72 & has\_belly\_color::yellow & 0.9978 & -0.2795 & 5.794\% \\
    96 & has\_leg\_color::grey & 0.9965 & -0.2105 & 4.723\% \\
    88 & has\_tail\_pattern::multi-colored & 0.9989 & -0.2308 & 4.306\% \\
    57 & has\_forehead\_color::black & 0.9999 & -0.2625 & 3.759\% \\
    99 & has\_bill\_color::grey & 0.9851 & -0.1078 & 3.269\% \\
    69 & has\_nape\_color::buff & 0.0001 & 0.2265 & 3.119\% \\
    24 & has\_breast\_pattern::multi-colored & 0.0026 & 0.1461 & 3.112\% \\
\end{longtable}
\end{center}

\begin{center}
    \begin{longtable}{|m{0.08\textwidth}|m{0.38\textwidth}|m{0.1\textwidth}|m{0.12\textwidth}|m{0.12\textwidth}|}
    \hline
    Concept ID &
    Concept name &  
    Concept Value &  
    Relevancy &  
    Concept contribution \\
    \hline\hline
    99 & has\_bill\_color::grey & 0.9999 & 0.0758 & 7.551\% \\
    88 & has\_tail\_pattern::multi-colored & 1.0 & 0.0691 & 6.883\% \\
    72 & has\_belly\_color::yellow & 1.0 & 0.0655 & 6.523\% \\
    18 & has\_underparts\_color::yellow & 1.0 & 0.0623 & 6.204\% \\
    110 & has\_wing\_pattern::striped & 1.0 & 0.0594 & 5.919\% \\
    47 & has\_throat\_color::black & 1.0 & 0.0525 & 5.227\% \\
    8 & has\_wing\_color::white & 0.9999 & 0.0517 & 5.153\% \\
\end{longtable}
\end{center}

\begin{center}
    \begin{longtable}{|m{0.08\textwidth}|m{0.38\textwidth}|m{0.1\textwidth}|m{0.12\textwidth}|m{0.12\textwidth}|}
    \hline
    Concept ID &
    Concept name &  
    Concept Value &  
    Relevancy &  
    Concept contribution \\
    \hline\hline
    81 & has\_shape::duck-like & 1.0 & 0.0799 & 7.399\% \\
    17 & has\_underparts\_color::grey & 1.0 & 0.0777 & 7.196\% \\
    98 & has\_leg\_color::buff & 1.0 & 0.0775 & 7.177\% \\
    45 & has\_throat\_color::grey & 1.0 & 0.0709 & 6.57\% \\
    40 & has\_breast\_color::grey & 1.0 & 0.068 & 6.296\% \\
    79 & has\_size::medium\_(9\_-\_16\_in) & 1.0 & 0.0613 & 5.678\% \\
    47 & has\_throat\_color::black & 1.0 & 0.053 & 4.905\% \\
\end{longtable}
\end{center}

\begin{center}
    \begin{longtable}{|m{0.08\textwidth}|m{0.38\textwidth}|m{0.1\textwidth}|m{0.12\textwidth}|m{0.12\textwidth}|}
    \hline
    Concept ID &
    Concept name &  
    Concept Value &  
    Relevancy &  
    Concept contribution \\
    \hline\hline
    79 & has\_size::medium\_(9\_-\_16\_in) & 1.0 & -0.1787 & 4.71\% \\
    47 & has\_throat\_color::black & 1.0 & -0.1923 & 4.628\% \\
    81 & has\_shape::duck-like & 1.0 & -0.2506 & 4.565\% \\
    17 & has\_underparts\_color::grey & 1.0 & -0.2023 & 4.072\% \\
    108 & has\_wing\_pattern::solid & 1.0 & -0.1343 & 3.805\% \\
    10 & has\_upperparts\_color::brown & 0.0 & 0.1386 & 3.356\% \\
    19 & has\_underparts\_color::black & 0.0 & 0.1474 & 3.227\% \\
\end{longtable}
\end{center}

\begin{center}
    \begin{longtable}{|m{0.08\textwidth}|m{0.38\textwidth}|m{0.1\textwidth}|m{0.12\textwidth}|m{0.12\textwidth}|}
    \hline
    Concept ID &
    Concept name &  
    Concept Value &  
    Relevancy &  
    Concept contribution \\
    \hline\hline
    45 & has\_throat\_color::grey & 0.9876 & -4.8754 & 8.199\% \\
    81 & has\_shape::duck-like & 0.9953 & -5.0468 & 6.939\% \\
    98 & has\_leg\_color::buff & 0.9697 & -2.2613 & 4.802\% \\
    79 & has\_size::medium\_(9\_-\_16\_in) & 1.0 & -5.745 & 4.032\% \\
    40 & has\_breast\_color::grey & 0.9263 & -1.3037 & 3.792\% \\
    52 & has\_bill\_length::shorter\_than\_head & 0.3725 & 0.2391 & 3.376\% \\
    61 & has\_under\_tail\_color::black & 0.9948 & -2.1669 & 3.033\% \\
\end{longtable}
\end{center}

\begin{center}
    \begin{longtable}{|m{0.08\textwidth}|m{0.38\textwidth}|m{0.1\textwidth}|m{0.12\textwidth}|m{0.12\textwidth}|}
    \hline
    Concept ID &
    Concept name &  
    Concept Value &  
    Relevancy &  
    Concept contribution \\
    \hline\hline
    81 & has\_shape::duck-like & 0.7868 & 0.0756 & 8.864\% \\
    45 & has\_throat\_color::grey & 0.8751 & 0.0763 & 8.052\% \\
    98 & has\_leg\_color::buff & 0.8302 & 0.0714 & 7.934\% \\
    40 & has\_breast\_color::grey & 0.9071 & 0.0691 & 7.028\% \\
    17 & has\_underparts\_color::grey & 0.8321 & 0.0592 & 6.563\% \\
    79 & has\_size::medium\_(9\_-\_16\_in) & 0.9978 & 0.0626 & 5.789\% \\
    42 & has\_breast\_color::black & 0.9994 & 0.0559 & 5.157\% \\
\end{longtable}
\end{center}

\begin{center}
    \begin{longtable}{|m{0.08\textwidth}|m{0.38\textwidth}|m{0.1\textwidth}|m{0.12\textwidth}|m{0.12\textwidth}|}
    \hline
    Concept ID &
    Concept name &  
    Concept Value &  
    Relevancy &  
    Concept contribution \\
    \hline\hline
    34 & has\_upper\_tail\_color::black & 0.2412 & 0.0209 & 6.27\% \\
    33 & has\_upper\_tail\_color::grey & 0.7204 & 0.0532 & 5.344\% \\
    35 & has\_upper\_tail\_color::white & 0.8736 & 0.0637 & 5.279\% \\
    38 & has\_head\_pattern::plain & 0.3886 & 0.0278 & 5.187\% \\
    58 & has\_forehead\_color::white & 0.953 & 0.0648 & 4.923\% \\
    107 & has\_crown\_color::white & 0.85 & 0.0542 & 4.615\% \\
    62 & has\_under\_tail\_color::white & 0.9583 & 0.0595 & 4.493\% \\
\end{longtable}
\end{center}

\begin{center}
    \begin{longtable}{|m{0.08\textwidth}|m{0.38\textwidth}|m{0.1\textwidth}|m{0.12\textwidth}|m{0.12\textwidth}|}
    \hline
    Concept ID &
    Concept name &  
    Concept Value &  
    Relevancy &  
    Concept contribution \\
    \hline\hline
    2 & has\_bill\_shape::all-purpose & 0.1927 & 0.0854 & 6.035\% \\
    35 & has\_upper\_tail\_color::white & 0.8736 & -0.0974 & 5.1\% \\
    62 & has\_under\_tail\_color::white & 0.9583 & -0.1473 & 4.753\% \\
    5 & has\_wing\_color::grey & 0.996 & -0.2048 & 3.753\% \\
    50 & has\_eye\_color::black & 0.9765 & -0.1364 & 3.705\% \\
    83 & has\_back\_pattern::solid & 0.9586 & -0.1122 & 3.614\% \\
    33 & has\_upper\_tail\_color::grey & 0.7204 & -0.0313 & 3.345\% \\
\end{longtable}
\end{center}

\begin{center}
    \begin{longtable}{|m{0.08\textwidth}|m{0.38\textwidth}|m{0.1\textwidth}|m{0.12\textwidth}|m{0.12\textwidth}|}
    \hline
    Concept ID &
    Concept name &  
    Concept Value &  
    Relevancy &  
    Concept contribution \\
    \hline\hline
    85 & has\_back\_pattern::multi-colored & 0.5726 & -0.0152 & 7.567\% \\
    1 & has\_bill\_shape::hooked\_seabird & 0.1999 & 0.0565 & 5.928\% \\
    58 & has\_forehead\_color::white & 0.6271 & -0.0124 & 3.467\% \\
    79 & has\_size::medium\_(9\_-\_16\_in) & 0.352 & 0.0144 & 3.438\% \\
    62 & has\_under\_tail\_color::white & 0.9619 & -0.0747 & 3.365\% \\
    91 & has\_primary\_color::grey & 0.943 & -0.0641 & 3.321\% \\
    111 & has\_wing\_pattern::multi-colored & 0.3383 & 0.0153 & 3.309\% \\
\end{longtable}
\end{center}

\begin{center}
    \begin{longtable}{|m{0.08\textwidth}|m{0.38\textwidth}|m{0.1\textwidth}|m{0.12\textwidth}|m{0.12\textwidth}|}
    \hline
    Concept ID &
    Concept name &  
    Concept Value &  
    Relevancy &  
    Concept contribution \\
    \hline\hline
    6 & has\_wing\_color::yellow & 0.2634 & 0.0367 & 7.529\% \\
    85 & has\_back\_pattern::multi-colored & 0.3738 & 0.0495 & 7.15\% \\
    12 & has\_upperparts\_color::yellow & 0.4082 & 0.0532 & 7.034\% \\
    104 & has\_crown\_color::grey & 0.9432 & 0.0994 & 5.689\% \\
    65 & has\_nape\_color::grey & 0.8918 & 0.0934 & 5.655\% \\
    111 & has\_wing\_pattern::multi-colored & 0.4636 & 0.0485 & 5.65\% \\
    55 & has\_forehead\_color::grey & 0.7109 & 0.0721 & 5.475\% \\
\end{longtable}
\end{center}

\begin{center}
    \begin{longtable}{|m{0.08\textwidth}|m{0.38\textwidth}|m{0.1\textwidth}|m{0.12\textwidth}|m{0.12\textwidth}|}
    \hline
    Concept ID &
    Concept name &  
    Concept Value &  
    Relevancy &  
    Concept contribution \\
    \hline\hline
    81 & has\_shape::duck-like & 0.9984 & 0.0741 & 5.731\% \\
    33 & has\_upper\_tail\_color::grey & 0.9931 & 0.0711 & 5.529\% \\
    79 & has\_size::medium\_(9\_-\_16\_in) & 0.9995 & 0.0674 & 5.207\% \\
    44 & has\_breast\_color::buff & 0.5648 & 0.0377 & 5.16\% \\
    111 & has\_wing\_pattern::multi-colored & 0.4532 & 0.03 & 5.117\% \\
    47 & has\_throat\_color::black & 0.6265 & 0.0414 & 5.106\% \\
    5 & has\_wing\_color::grey & 0.9933 & 0.0655 & 5.095\% \\
\end{longtable}
\end{center}

\begin{center}
    \begin{longtable}{|m{0.08\textwidth}|m{0.38\textwidth}|m{0.1\textwidth}|m{0.12\textwidth}|m{0.12\textwidth}|}
    \hline
    Concept ID &
    Concept name &  
    Concept Value &  
    Relevancy &  
    Concept contribution \\
    \hline\hline
    81 & has\_shape::duck-like & 0.9984 & -0.1687 & 4.873\% \\
    28 & has\_back\_color::black & 0.9058 & -0.0491 & 4.025\% \\
    79 & has\_size::medium\_(9\_-\_16\_in) & 0.9995 & -0.1617 & 3.938\% \\
    20 & has\_underparts\_color::white & 0.9081 & -0.0484 & 3.919\% \\
    111 & has\_wing\_pattern::multi-colored & 0.4532 & 0.0039 & 3.843\% \\
    33 & has\_upper\_tail\_color::grey & 0.9931 & -0.0961 & 3.587\% \\
    67 & has\_nape\_color::black & 0.5885 & -0.0068 & 3.519\% \\
\end{longtable}
\end{center}

\begin{center}
    \begin{longtable}{|m{0.08\textwidth}|m{0.38\textwidth}|m{0.1\textwidth}|m{0.12\textwidth}|m{0.12\textwidth}|}
    \hline
    Concept ID &
    Concept name &  
    Concept Value &  
    Relevancy &  
    Concept contribution \\
    \hline\hline
    81 & has\_shape::duck-like & 1.0 & 2.2956 & 5.997\% \\
    33 & has\_upper\_tail\_color::grey & 0.9984 & 0.8142 & 4.54\% \\
    44 & has\_breast\_color::buff & 0.7583 & 0.1373 & 4.311\% \\
    28 & has\_back\_color::black & 0.9992 & 0.7695 & 3.871\% \\
    111 & has\_wing\_pattern::multi-colored & 0.91 & 0.2389 & 3.706\% \\
    79 & has\_size::medium\_(9\_-\_16\_in) & 1.0 & 1.1385 & 3.602\% \\
    14 & has\_upperparts\_color::white & 0.9258 & 0.2519 & 3.583\% \\
\end{longtable}
\end{center}

\begin{center}
    \begin{longtable}{|m{0.08\textwidth}|m{0.38\textwidth}|m{0.1\textwidth}|m{0.12\textwidth}|m{0.12\textwidth}|}
    \hline
    Concept ID &
    Concept name &  
    Concept Value &  
    Relevancy &  
    Concept contribution \\
    \hline\hline
    81 & has\_shape::duck-like & 1.0 & 0.1233 & 6.881\% \\
    79 & has\_size::medium\_(9\_-\_16\_in) & 1.0 & 0.112 & 6.251\% \\
    69 & has\_nape\_color::buff & 0.2768 & 0.0266 & 5.36\% \\
    49 & has\_throat\_color::buff & 0.1734 & 0.0161 & 5.188\% \\
    39 & has\_breast\_color::brown & 0.3031 & 0.0282 & 5.185\% \\
    60 & has\_under\_tail\_color::grey & 0.9929 & 0.0905 & 5.087\% \\
    33 & has\_upper\_tail\_color::grey & 0.9996 & 0.0885 & 4.943\% \\
\end{longtable}
\end{center}

\twocolumn
\end{document}